\documentclass[%
 aip,
 amsmath,amssymb,
 reprint,%
]{revtex4-1}
\usepackage{array}
\usepackage{graphicx}
\usepackage{dcolumn}
\usepackage{bm}
\usepackage{amsmath}
\usepackage[utf8]{inputenc}
\usepackage[T1]{fontenc}
\usepackage{mathptmx}
\usepackage{etoolbox}
\usepackage{booktabs}
\usepackage[colorlinks=true, linkcolor=blue, citecolor=blue, urlcolor=blue ]{hyperref}
\makeatletter
\def\@email#1#2{%
 \endgroup
 \patchcmd{\titleblock@produce}
  {\frontmatter@RRAPformat}
  {\frontmatter@RRAPformat{\produce@RRAP{*#1\href{mailto:#2}{#2}}}\frontmatter@RRAPformat}
  {}{}
}%
\makeatother
\begin{document}

\preprint{AIP/123-QED}

\title[Sample title]{Can Deep Learning Achieve Cross-Physics Mapping?}
\author{Pengfei Zhu*}
\author{Julien Lecompagnon}%
\author{Mathias Ziegler}
 \email{pengfei.zhu@bam.de}
\affiliation{ 
Bundesanstalt für Materialforschung and -prüfung (BAM), 12205 Berlin, Germany
}%

\date{\today}

\begin{abstract}
Can deep learning translate physical fields governed by fundamentally different equations? We address this question by introducing Cross-Physics Mapping (CPM), an operator-learning framework for mappings between heterogeneous physical domains. We formulate sufficient conditions for such mappings through compatible latent representations and propose a dimensionless scaling principle that aligns the characteristic evolution scales of the source and target systems without assuming their dynamical equivalence. As a representative test, paired diffusion and wave fields are generated independently from their respective parabolic and hyperbolic equations while sharing the same latent geometry, material heterogeneity, excitation, and dimensionless scale. Seven architectures—ResUNet, DeepONet, Fourier, latent, wavelet, U-shaped, and Galerkin neural operators—are evaluated for both diffusion-to-wave and wave-to-diffusion mappings. The results reveal a strong directional asymmetry. Diffusion-to-wave reconstruction is more challenging because it requires recovering wavefront, phase, and time-of-flight information attenuated by diffusion; U-NO performs best in this direction, achieving a relative $\ell_2$ error of $0.307$ and an $R^2$ of $0.905$. Wave-to-diffusion mapping is considerably more stable, with GNO attaining a relative $\ell_2$ error of $0.154$ and an $R^2$ of $0.935$. Neural operators generally outperform the conventional convolutional baseline, highlighting the nonlocal nature of cross-physics transformations. These findings demonstrate that deep learning can establish useful mappings between distinct physical modalities on a shared latent manifold, while the achievable accuracy remains fundamentally constrained by the direction-dependent information content of the governing physics.
\end{abstract}

\maketitle

\section{Introduction}
The integration of machine learning with physical modeling has emerged as a transformative paradigm for scientific computing \cite{carleo2019machine, karniadakis2021physics, karagiorgi2022machine}. Traditional numerical methods, including finite difference \cite{zhu2023characterization}, finite element \cite{jiang2024non}, and spectral approaches \cite{stoll2006easyspin}, have achieved remarkable success in solving forward and inverse problems governed by partial differential equations (PDEs) \cite{zhu2025making,zhu2026thermal}. However, these methods often require expensive mesh generation, repeated numerical simulations, and substantial computational resources, especially when dealing with nonlinear \cite{kundert1986simulation}, multiphysics \cite{zhu2024novel}, and multiscale systems \cite{zeng2008multiscale}.

To address these challenges, Physics-Informed Neural Networks (PINNs) were introduced as a framework that embeds governing physical laws directly into the training process of neural networks \cite{raissi2018physics,cai2021physics,cai2021physics2}. Instead of relying solely on observational data, PINNs incorporate PDE residuals, boundary conditions, and initial conditions into the loss function, enabling neural networks to learn physically consistent solutions with limited data.   Since their introduction, PINNs have been successfully applied to a broad range of forward and inverse problems, including fluid dynamics \cite{raissi2020hidden}, wave propagation \cite{zhu2025modeling}, quantum mechanics \cite{norambuena2024physics}, electromagnetics \cite{zhu2026thz}, and material characterization \cite{zhu2025thz,zhu2025physics}.
Despite their success, PINNs primarily focus on approximating solutions of a given physical system. To improve scalability and representation capability, several extensions have been proposed, including variational PINNs (VPINNs) \cite{kharazmi2019variational}, hp-VPINNs \cite{kharazmi2021hp}, and XPINNs \cite{jagtap2020extended}, which introduce variational formulations, domain decomposition, and adaptive refinement strategies. These developments significantly expanded the applicability of physics-informed learning but remained largely centered on solving or identifying individual PDE systems.

\begin{figure*}[t]
	\includegraphics[width=\textwidth]{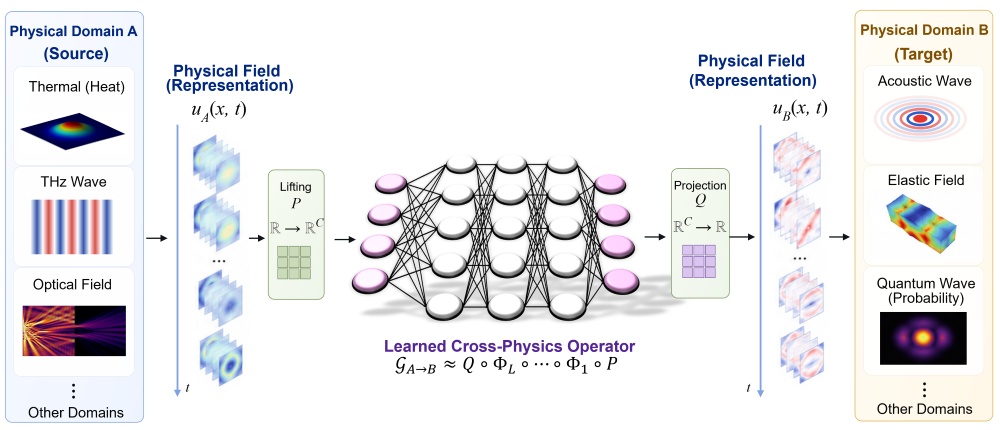}
	\caption{Conceptual illustration of Cross-Physics Mapping (CPM). A physical field $u_A(x,t)$ from a source domain is mapped to a field $u_B(x,t)$ in a different target domain through a learned operator $\mathcal{G}_{A\rightarrow B}$. The source and target domains may correspond to distinct physical phenomena, including thermal, optical, acoustic, elastic, or quantum systems. The mapping is formulated at the function-space level and can be approximated by different learning architectures.}
	\label{fig1}
\end{figure*}

A parallel development has emerged in the form of neural operators \cite{azizzadenesheli2024neural,raonic2023convolutional,kovachki2023neural}. Unlike conventional neural networks that learn mappings between finite-dimensional vectors, neural operators learn mappings between function spaces and directly approximate solution operators of PDE families. Representative examples include DeepONet \cite{lu2019deeponet} and the Fourier Neural Operator (FNO) \cite{li2020fourier}, which demonstrated the ability to learn entire classes of physical systems and perform zero-shot generalization across unseen inputs and discretizations \cite{zhu2026principal,zhu2025real}. These advances shifted the focus from learning individual solutions toward learning underlying physical transformations.
However, most existing operator-learning frameworks remain restricted to mappings within the same physical domain. Typical examples include learning the evolution of fluid fields \cite{toscano2026mr}, predicting electromagnetic responses from material parameters \cite{zhu2026field}, or approximating PDE solution operators \cite{de2022generic}. In these settings, both input and output belong to the same governing physics, differing only in parameters, boundary conditions, or temporal evolution. As a result, current neural operators predominantly learn intra-physics transformations rather than relationships between fundamentally different physical modalities.

Many practical sensing and imaging systems, however, inherently involve interactions between distinct physical domains \cite{zhu2025comprehensive,zhu2025frequency}. For instance, thermal diffusion generates infrared signals \cite{zhu2026novel}, acoustic waves induce mechanical vibrations \cite{xiao2022bioinspired}, electromagnetic excitations produce thermal responses \cite{1132767}, and terahertz scattering encodes material properties through wave–matter interactions \cite{}. These processes suggest the existence of latent operators that connect heterogeneous physical fields. Recovering such operators would enable the translation of information from one modality to another, potentially revealing representations that are more informative, more robust, or easier to interpret than the original measurements.
From a signal and image processing perspective, different physical modalities can be viewed as distinct representations of the same underlying object or scene. Each sensing mechanism acts as a physics-dependent transform that selectively preserves, suppresses, or redistributes information across spatial, temporal, and spectral dimensions. For example, thermal diffusion behaves as a strong low-pass operator that smooths high-frequency details, whereas wave-based modalities preserve phase and propagation information. Similarly, optical, acoustic, and electromagnetic measurements emphasize different aspects of the same physical structure through their respective transfer functions. Consequently, information that is difficult to extract in one modality may become significantly more accessible in another.

Motivated by this observation, we introduce the concept of "Cross-Physics Mapping (CPM)", as shown in Fig.~\ref{fig1}. Unlike conventional operator learning that focuses on solution operators within a single governing equation, CPM aims to learn operators between distinct physical domains. Formally, CPM seeks mappings of the form
\begin{equation}
	\mathcal{G}: \mathcal{F}*{A} \rightarrow \mathcal{F}*{B},
\end{equation}
where ($\mathcal{F}*{A}$) and ($\mathcal{F}*{B}$) denote function spaces associated with different physical phenomena. Examples include thermal-to-wave, thermal-to-acoustic, diffusion-to-electromagnetic, and infrared-to-terahertz transformations. Such mappings are not intended to replace the underlying physics but rather to uncover physically meaningful latent representations that bridge different modalities.
This perspective extends physics-informed learning from solving equations to connecting physical worlds. By treating heterogeneous physical fields as related operator spaces, Cross-Physics Mapping establishes a new framework that unifies sensing, imaging, inverse problems, and scientific machine learning under a common operator-learning parTadigm. It opens opportunities for physics-guided modality translation, enhanced information recovery, and the discovery of hidden relationships across seemingly unrelated physical processes. Models and codes are available at \url{https://github.com/Pengfei0705/Cross-Physics-Learning}.

\paragraph*{Theorem 1 (Cross-Physics Operator Mapping)}
\textit{Let $\mathcal{F}_A$ and $\mathcal{F}_B$ be two function spaces associated with physical fields
	$u_A$ and $u_B$, governed respectively by
	$
	\mathcal{P}_A(u_A;\theta_A)=0,
	\mathcal{P}_B(u_B;\theta_B)=0,
	$
	where $\mathcal{P}_A$ and $\mathcal{P}_B$ are possibly different physical operators, and
	$\theta_A,\theta_B$ denote the corresponding physical parameters.
	Assume that both systems admit latent operator representations
	$
	u_A = \Phi_A z_A,
	u_B = \Phi_B z_B,
	$
	where $\Phi_A$ and $\Phi_B$ are representation operators and
	$z_A \in \mathcal{Z}_A$, $z_B \in \mathcal{Z}_B$
	are latent coordinates.
	If there exists a bounded latent transformation
	$
	\mathcal{M}_{A\rightarrow B}:
	\mathcal{Z}_A
	\rightarrow
	\mathcal{Z}_B
	$
	such that
	$
	z_B
	=
	\mathcal{M}_{A\rightarrow B}(z_A),
	$
	then there exists a cross-physics operator
	$
	\mathcal{G}_{A\rightarrow B}:
	\mathcal{F}_A
	\rightarrow
	\mathcal{F}_B
	$
	satisfying}
\begin{equation}
	u_B
	=
	\mathcal{G}_{A\rightarrow B}(u_A)
	=
	\Phi_B
	\mathcal{M}_{A\rightarrow B}
	\Phi_A^{-1}
	u_A.
\end{equation}
\textit{Furthermore, if $\mathcal{M}_{A\rightarrow B}$ is continuous on a compact subset
	of $\mathcal{Z}_A$, then for every $\varepsilon>0$ there exists a neural operator
	$\mathcal{G}_{\Theta}$ such that}
\begin{equation}
	\sup_{u_A\in\mathcal{K}}
	\left\|
	\mathcal{G}_{A\rightarrow B}(u_A)
	-
	\mathcal{G}_{\Theta}(u_A)
	\right\|_{\mathcal{F}_B}
	<
	\varepsilon,
\end{equation}
\textit{for any compact set $\mathcal{K}\subset\mathcal{F}_A$.}

The proof follows from the existence of a common latent representation.
By assumption,
$
u_A = \Phi_A z_A,
u_B = \Phi_B z_B.
$
Since the latent variables are related through
$
z_B
=
\mathcal{M}_{A\rightarrow B}(z_A),
$,
we obtain
$
u_B
=
\Phi_B
\mathcal{M}_{A\rightarrow B}(z_A).
$.
Assuming that $\Phi_A$ is injective or admits a stable pseudo-inverse on the
relevant data manifold,
$
z_A
=
\Phi_A^{-1}u_A.
$.
Substituting this relation into the previous equation yields
\begin{equation}
	u_B
	=
	\Phi_B
	\mathcal{M}_{A\rightarrow B}
	\Phi_A^{-1}
	u_A.
\end{equation}

Defining
$
\mathcal{G}_{A\rightarrow B}
=
\Phi_B
\circ
\mathcal{M}_{A\rightarrow B}
\circ
\Phi_A^{-1},
$
we obtain
$
u_B
=
\mathcal{G}_{A\rightarrow B}(u_A),
$
which establishes the existence of a cross-physics operator.
Since $\mathcal{M}_{A\rightarrow B}$ is continuous on a compact domain,
$\mathcal{G}_{A\rightarrow B}$ is also continuous.
By the universal approximation theorem for neural operators,
there exists a neural operator $\mathcal{G}_{\Theta}$ satisfying
\begin{equation}
	\sup_{u_A\in\mathcal{K}}
	\left\|
	\mathcal{G}_{A\rightarrow B}(u_A)
	-
	\mathcal{G}_{\Theta}(u_A)
	\right\|_{\mathcal{F}_B}
	<
	\varepsilon.
\end{equation}

Therefore, the cross-physics mapping can be approximated arbitrarily well by a neural operator.
The theorem does not require the two physical systems to obey the same governing
equation, nor does it assume dynamical equivalence. It only requires the existence
of compatible latent operator representations. Consequently, thermal-to-acoustic,
diffusion-to-wave, electromagnetic-to-thermal, optical-to-mechanical,
and other heterogeneous physical transformations can all be interpreted as
special instances of Cross-Physics Mapping.

\begin{figure*}[t]
	\includegraphics[width=\textwidth]{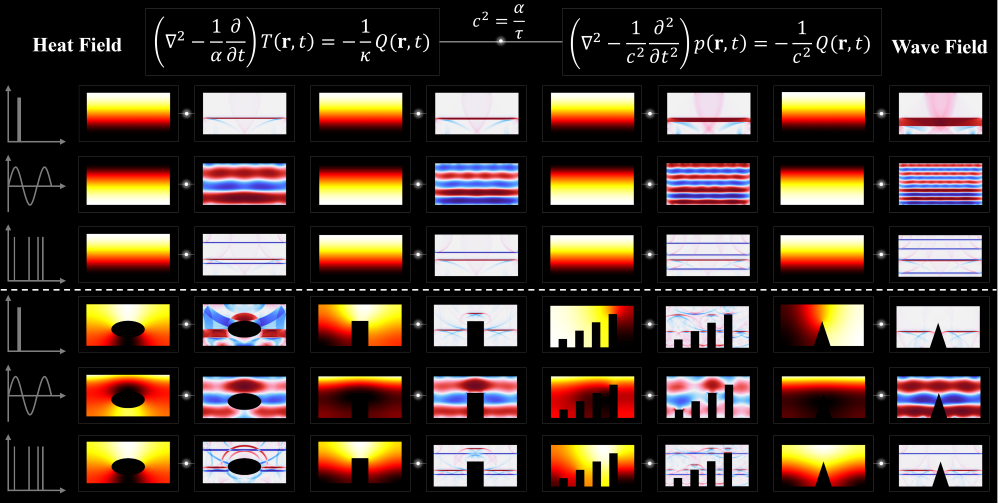}
	\caption{COMSOL-generated paired dataset for cross-physics mapping between diffusive and wave fields.
		Each sample consists of a heat-field response $T(\mathbf{r},t)$ governed by
		$\left(\nabla^2-\alpha^{-1}\partial_t\right)T=-Q/\kappa$
		and its corresponding wave-field response $p(\mathbf{r},t)$ governed by
		$\left(\nabla^2-c^{-2}\partial_t^2\right)p=-Q/c^2$, with the characteristic relation
		$c^2=\alpha/\tau$.
		The upper three rows illustrate paired responses under representative temporal excitations, including impulsive, harmonic, and pulse-train sources, while the lower three rows show examples with spatially heterogeneous geometries and obstacles.
		For each configuration, the same source and material geometry are used to generate the diffusion--wave pair, providing aligned input--output samples for learning the cross-physics operator.
		The dataset spans variations in excitation waveform, geometry, and spatial structure, enabling the model to learn a general mapping from diffusion-dominated thermal fields to their propagation-dominated wave counterparts.
	}
	\label{fig2}
\end{figure*}

\section{Cross-Physics Scaling Principle}
Although different physical domains may be governed by
distinct equations and material parameters, their dynamics
can often be characterized through a small number of
dimensionless operator groups.
From the perspective of operator learning, these groups
determine the intrinsic spatial, temporal, and spectral
scales of the underlying physical process.

Consider a physical field governed by
$
\mathcal P(u;\theta)=0,
$
where $\theta$ denotes the associated material parameters.
Introducing characteristic spatial and temporal scales,
$L$ and $t_0$, yields the dimensionless form
\begin{equation}
	\mathcal P^*(u^*;\Pi_1,\Pi_2,\ldots)=0,
\end{equation}
where $\Pi_i$ denote the dominant dimensionless groups.
Examples include the Fourier number for diffusion,
the propagation parameter for wave dynamics,
and analogous quantities in electromagnetic,
and quantum systems.
The central assumption of this work is that cross-physics
mappings are most stable when the dominant dimensionless
operator scales are comparable across domains.
Therefore, for a mapping
$
\mathcal G_{A\rightarrow B}:
\mathcal F_A
\rightarrow
\mathcal F_B,
$
we seek source and target systems satisfying
$
\Pi_A
\sim
\Pi_B.
$
This criterion does not imply dynamical equivalence between
the two physical systems. Rather, it ensures that the
corresponding fields evolve on compatible regions of the
underlying operator manifold, thereby reducing the scale
mismatch that must be learned by the neural operator.
As a simple example, matching the dominant scales of a
diffusive process and a wave process yields
$
\mathrm{Fo}
\sim
\lambda^2,
$
which leads to
$
c^2
\sim
\alpha/t_0.
$
The diffusion-to-wave correspondence is therefore recovered
as a special case of the proposed scaling framework.
Detailed derivations for diffusion, wave, electromagnetic,
and quantum systems are provided in
Supplementary Material A.

\begin{figure}[t]
\includegraphics[width=0.5\textwidth]{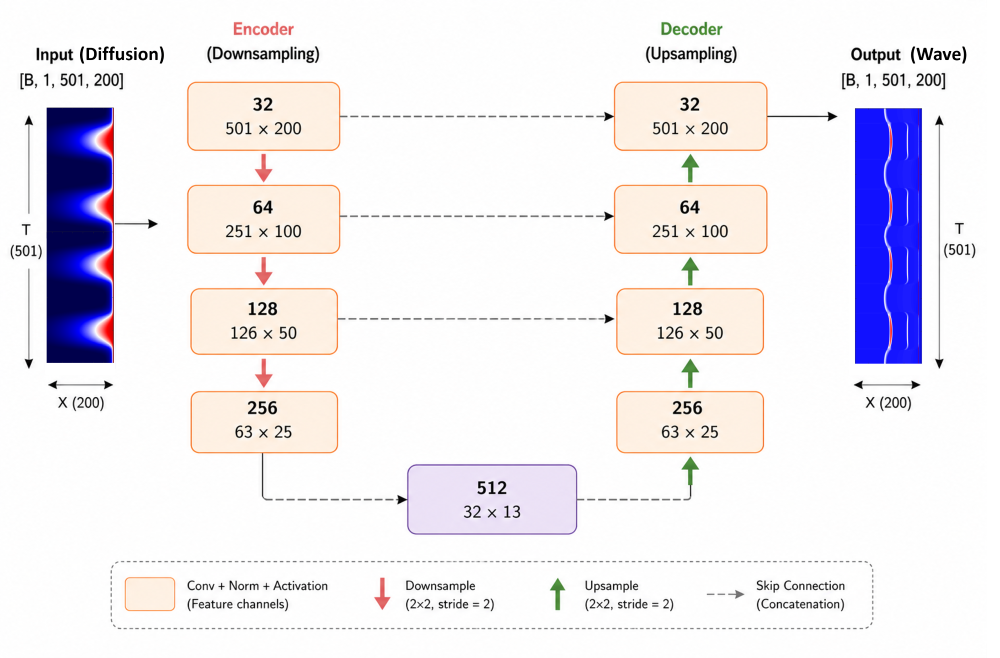}
\caption{ResUNet architecture for diffusion-to-wave cross-physics mapping.
	The network maps a spatiotemporal diffusion field
	$T(x,t)\in\mathbb{R}^{501\times200}$
	to the corresponding wave field
	$p(x,t)\in\mathbb{R}^{501\times200}$.
	The encoder progressively extracts multiscale representations through
	convolutional residual blocks and $2\times2$ downsampling, increasing the
	feature dimension from 32 to 512 channels while reducing the spatiotemporal
	resolution.
	The decoder symmetrically restores the original resolution through successive
	upsampling stages.
	Skip connections concatenate encoder features with the corresponding decoder
	features at matched scales, preserving fine-scale spatiotemporal information
	that may otherwise be lost in the latent representation.
	The bottleneck provides a compact high-dimensional representation connecting
	the diffusive and propagative physical domains.
	Numbers inside each block indicate the number of feature channels and the
	corresponding $(T\times X)$ feature-map resolution.}
\label{fig3}
\end{figure}

The scaling principle provides a practical guideline for
constructing cross-physics datasets and selecting simulation
parameters.
Given a source domain and a target domain, the characteristic
length scale, observation window, and material parameters
should be chosen such that their dominant dimensionless
groups remain comparable.
This requirement also determines the spectral complexity of
the learned operator. Physical processes with larger
characteristic bandwidths require higher spatial resolution
and more retained Fourier modes, whereas strongly dissipative
systems naturally admit lower-dimensional spectral
representations. Consequently, the Fourier truncation level
in the neural operator is not merely a numerical
hyperparameter, but a reflection of the physical scales
present in the mapped domains.
When the dimensionless scales are matched, the source and
target fields occupy compatible regions of the latent
operator manifold, making the learned cross-physics mapping
more stable, data-efficient, and physically interpretable.

\section{Data Generation for Cross-Physics Operator Learning}
\label{sec:data_generation}
A central requirement of cross-physics operator learning is that the input and target fields must correspond to the same underlying structural configuration, while being generated by different governing equations. We therefore construct a paired synthetic dataset in which each sample is defined by a common latent physical scene---including geometry, subsurface heterogeneity, excitation location, and characteristic scales---and is propagated independently through a diffusion model and a wave model. This construction provides paired observations of the form
$
\mathcal{D}
=
\left\{
\left(
u_{\mathrm{D}}^{(n)},
u_{\mathrm{W}}^{(n)}
\right)
\right\}_{n=1}^{N_s},
$
where $u_{\mathrm{D}}$ denotes the diffusion-domain observation and
$u_{\mathrm{W}}$ denotes the corresponding wave-domain observation.
Importantly, we do not assume that the two fields are related by a pointwise or closed-form transformation. Instead, both fields are conditioned on the same latent structural realization. The learning problem is therefore to identify the nonlinear operator
$
\mathcal{G}:
u_{\mathrm{D}}
\mapsto
u_{\mathrm{W}},
$
from paired observations generated under matched structural and dimensionless conditions.

\begin{figure}[t]
	\includegraphics[width=0.5\textwidth]{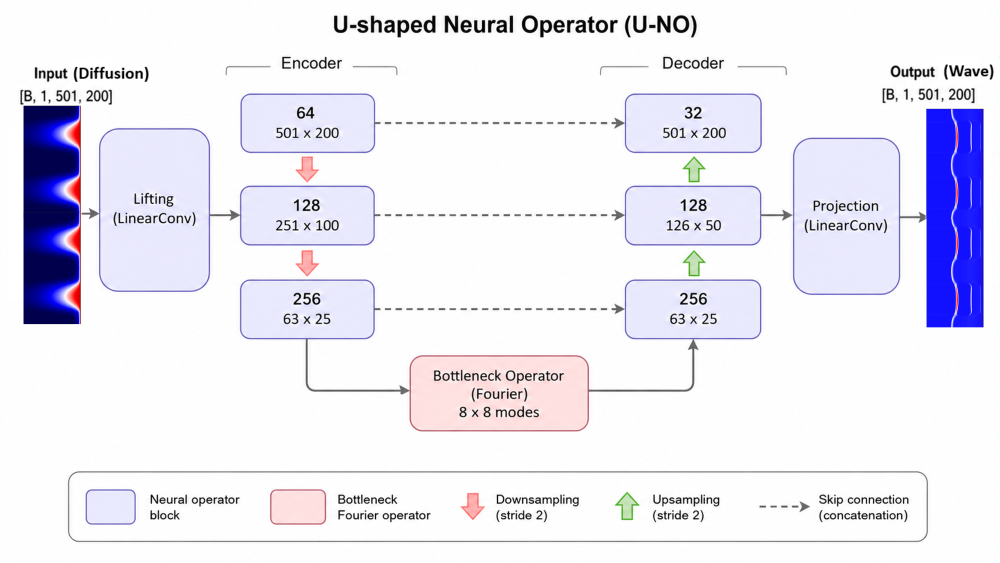}
	\caption{U-Shaped Neural Operator (U-NO) for diffusion-to-wave cross-physics mapping.
		The model learns the operator mapping from an input diffusion field
		$T(x,t)\in\mathbb{R}^{501\times200}$
		to the corresponding wave field
		$p(x,t)\in\mathbb{R}^{501\times200}$.
		A lifting layer first embeds the scalar input field into a higher-dimensional
		feature representation, which is progressively compressed by the U-shaped
		encoder to construct multiscale latent features.
		At the coarsest representation, a Fourier neural operator with $8\times8$
		retained modes captures nonlocal interactions in the joint space--time domain.
		The decoder subsequently restores the original spatiotemporal resolution,
		while skip connections transfer multiscale encoder features to the
		corresponding decoder stages.
		Finally, a linear projection maps the decoded representation back to the
		physical wave field.
		Numbers inside the blocks denote feature channels and $(T\times X)$
		feature-map resolutions, respectively.}
	\label{fig4}
\end{figure}

\subsection{Computational domain and latent structural model}
The physical fields are simulated on a two-dimensional rectangular domain
$
\Omega
=
[0,L_x]
\times
[0,L_y],
$
with
$
L_x=0.2,
L_y=1.0.
$
The numerical domain is discretized using
$
N_x\times N_y
=
256\times96
$
grid points. The corresponding grid spacings are
$
\Delta x
=
L_x/(N_x-1),
\Delta y
=
L_y/(N_y-1).
$
For each realization, a smooth subsurface inclusion is generated to represent a localized material heterogeneity. Its geometry is parameterized by
$
\boldsymbol{\theta}_d
=
\left(
x_c,
y_c,
a,
b,
\psi,
\gamma
\right),
$
where $(x_c,y_c)$ denotes the inclusion center, $a$ and $b$ are the semi-major and semi-minor axes, $\psi$ is the orientation angle, and $\gamma$ controls the material contrast.
Defining the rotated coordinates
$
x' =
(x-x_c)\cos\psi
+
(y-y_c)\sin\psi,
y' =
-(x-x_c)\sin\psi
+
(y-y_c)\cos\psi,
$
the inclusion level-set function is
$
r_d(x,y)
=
\left(
x'/a
\right)^2
+
\left(
y'/b
\right)^2.
$
A smooth material map is constructed as
\begin{equation}
	m(x,y)
	=
	\frac{1}{2}
	\left[
	1
	-
	\tanh
	\left(
	\frac{
		r_d(x,y)-1
	}{
		\varepsilon_d
	}
	\right)
	\right],
	\label{eq:defect_map}
\end{equation}
where $\varepsilon_d$ controls the transition width at the inclusion boundary. Thus,
$m\approx1$ inside the inclusion and $m\approx0$ in the background.
Rather than using the defect map only as an auxiliary label, we explicitly incorporate it into the governing coefficients. The diffusion and wave parameters are defined as
$
\alpha(x,y)
=
\alpha_b
\left[
1+
\delta_{\alpha}
m(x,y)
\right],
$
and
$
c(x,y)
=
c_b
\left[
1+
\delta_c
m(x,y)
\right],
$
where $\alpha_b$ and $c_b$ denote the background diffusivity and wave speed, while $\delta_\alpha$ and $\delta_c$ control the corresponding material contrasts. Consequently, both physical fields are influenced by the same latent structure, but through different governing operators.

Both simulations are driven from the accessible surface
$
\Gamma_s
=
\left\{
(x,y)\in\Omega:
y=0
\right\}.
$
For each sample, we excited different excitations $g_x(x)$ including amplitude, location, type (finite-width pulse, chirped pulse, and lock-in).

\begin{figure}[t]
	\includegraphics[width=0.5\textwidth]{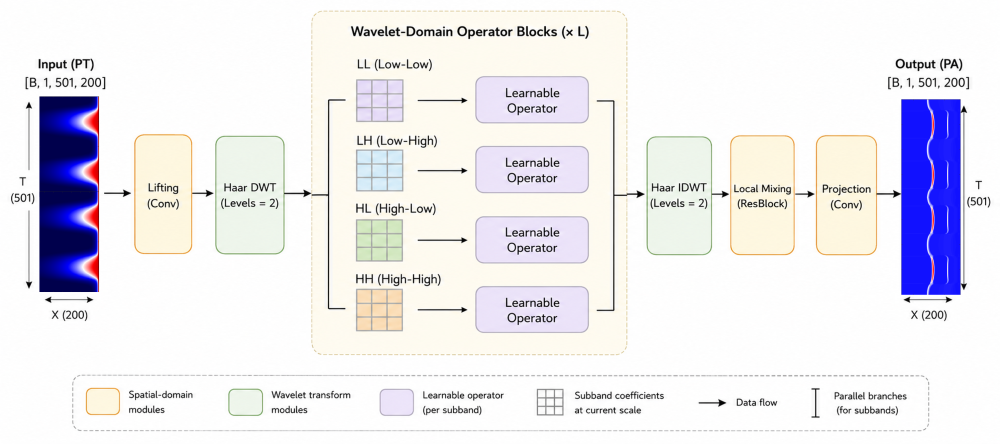}
	\caption{Wavelet Neural Operator (WNO) for cross-physics operator learning.
		Given an input diffusion field
		$T(x,t)\in\mathbb{R}^{501\times200}$,
		WNO predicts the corresponding wave field
		$p(x,t)\in\mathbb{R}^{501\times200}$
		through a multiresolution wavelet representation.
		The input is first lifted to a higher-dimensional feature space and decomposed
		using a two-level Haar discrete wavelet transform (DWT).
		The resulting low--low (LL), low--high (LH), high--low (HL), and high--high
		(HH) subbands are processed by parallel learnable operator branches,
		allowing the model to capture cross-physics interactions at distinct
		spatiotemporal scales and frequency bands.
		After $L$ wavelet-domain operator blocks, the transformed features are
		reconstructed by the inverse discrete wavelet transform (IDWT).
		A residual local-mixing block further refines short-range features before a
		final projection maps the representation to the physical wave field.
		This multiresolution construction provides an explicit mechanism for jointly
		representing smooth diffusive structures and rapidly varying propagative
		features within the learned diffusion-to-wave operator.}
	\label{fig5}
\end{figure}

\subsection{Diffusion-domain simulation}
The source-domain field is modeled by the heterogeneous diffusion equation
\begin{equation}
	\frac{\partial u}{\partial t}
	=
	\nabla\cdot
	\left[
	\alpha(x,y)
	\nabla u
	\right]
	+
	s_d(x,y,t),
	\qquad
	(x,y)\in\Omega,
	\label{eq:diffusion_pde}
\end{equation}
where $u(x,y,t)$ denotes the diffusion field and $s_d$ represents a possible volumetric source contribution.
For surface excitation, Eq.~\eqref{eq:diffusion_pde} is combined with a prescribed boundary flux,
$
-
\mathbf{n}\cdot
\left[
\alpha(x,y)
\nabla u
\right]
=
q_d(x,t),
(x,y)\in\Gamma_s,
$
where $\mathbf{n}$ is the outward unit normal. The remaining boundaries are taken to be homogeneous during the observation window,
$
\mathbf{n}\cdot\nabla u
=
0,
(x,y)\in
\partial\Omega\setminus\Gamma_s,
$
and the initial condition is
$
u(x,y,0)=0.
$
The diffusion field can be written formally as
$
u(\cdot,t)
=
\mathcal{S}_{d}
\left[
m,
\alpha,
q_d
\right],
$
where $\mathcal{S}_{d}$ denotes the numerical solution operator associated with the diffusion equation.
The PDE is discretized using second-order spatial finite differences. For an explicit time-marching implementation, the temporal step is selected to satisfy the diffusion stability condition
\begin{equation}
	\Delta t_d
	\leq
	\frac{1}{
		2\alpha_{\max}
		\left(
		\Delta x^{-2}
		+
		\Delta y^{-2}
		\right)
	},
	\label{eq:diffusion_stability}
\end{equation}
where
$
\alpha_{\max}
=
\max_{(x,y)\in\Omega}
\alpha(x,y).
$
The solver therefore operates at an internal temporal resolution sufficiently fine to accurately resolve the diffusion dynamics.

\begin{figure}[t]
	\includegraphics[width=0.5\textwidth]{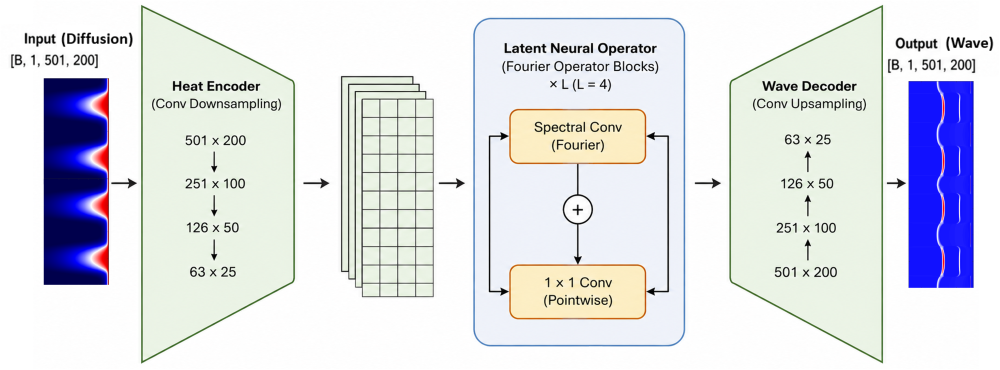}
	\caption{Latent Neural Operator (LNO) for cross-physics operator learning.
		Given an input diffusion field
		$T(x,t)\in\mathbb{R}^{501\times200}$,
		LNO predicts the corresponding wave field
		$p(x,t)\in\mathbb{R}^{501\times200}$
		by learning the cross-physics transformation in a compressed latent space.
		A convolutional heat encoder progressively downsamples the input field from
		$501\times200$ to $63\times25$, producing a compact multichannel latent
		representation.
		The latent field is then processed by $L=4$ Fourier operator blocks, each
		combining a spectral convolution for nonlocal interactions with a pointwise
		$1\times1$ convolution for local feature mixing.
		The resulting latent representation is mapped back to the physical domain by
		a convolutional wave decoder that progressively restores the original
		spatiotemporal resolution.
		By performing operator learning on the compressed representation, LNO provides
		an efficient mechanism for capturing global diffusion-to-wave dependencies
		without applying spectral operations directly at the full field resolution.}
	\label{fig6}
\end{figure}

\subsection{Wave-domain simulation}
The target field is generated independently from the corresponding heterogeneous wave equation,
\begin{equation}
	\frac{\partial^2 v}{\partial t^2}
	=
	\nabla\cdot
	\left[
	c^2(x,y)
	\nabla v
	\right]
	+
	s_w(x,y,t),
	\qquad
	(x,y)\in\Omega,
	\label{eq:wave_pde}
\end{equation}
where $v(x,y,t)$ denotes the wave-domain field.
The initial conditions are
$
v(x,y,0)=0,
\partial v(x,y,0)/\partial t=0.
$
The excitation is applied on the same accessible surface $\Gamma_s$. The resulting solution can be represented as
$
v(\cdot,t)
=
\mathcal{S}_{w}
\left[
m,
c,
q_w
\right],
$
where $\mathcal{S}_{w}$ is the wave propagation solution operator.
The wave equation is discretized using second-order finite differences in both space and time. The temporal step satisfies the two-dimensional Courant--Friedrichs--Lewy condition
\begin{equation}
	c_{\max}\Delta t_w
	\sqrt{
		\frac{1}{\Delta x^2}
		+
		\frac{1}{\Delta y^2}
	}
	\leq
	C_{\mathrm{CFL}},
	\label{eq:wave_cfl}
\end{equation}
where $c_{\max}$ is the maximum wave speed in the domain and
$C_{\mathrm{CFL}}\leq1$ is the selected stability factor.
In contrast to the diffusion field, whose high-spatial-frequency components are progressively attenuated, the wave field retains oscillatory phase information and finite-speed propagation. The mapping between the two domains is therefore nontrivial and cannot, in general, be reduced to a local amplitude transformation.

\subsection{Dimensionless cross-physics matching}
Because diffusion and wave propagation possess fundamentally different characteristic time scales, directly sampling $\alpha$, $c$, and the observation time independently can produce physically incomparable pairs. We therefore control the simulations through dimensionless characteristic groups.
For a characteristic length $L$ and time $t_0$, the diffusion dynamics are characterized by the Fourier number
$
\mathrm{Fo}
=
\alpha_b t_0
/
L^2,
$
whereas wave propagation is characterized by the dimensionless propagation distance
$
\Lambda
=
c_b t_0
/
L.
$
To generate paired fields at comparable dimensionless evolution scales, we impose the similarity condition
$
\mathrm{Fo}
=
\Lambda^2.
$
Substitution of the Fourier number and the dimensionless wave number gives
$
\alpha_b t_0
/
L^2
=
c_b^2t_0^2
/
L^2,
$
or equivalently,
$
\alpha_b
=
c_b^2t_0.
$
The above equation should not be interpreted as an analytical equivalence between the diffusion and wave equations. Rather, it is used as a data-generation constraint that prevents the two simulations from occupying arbitrarily different characteristic scales. It ensures that the characteristic diffusion length
$
\ell_d
\sim
\sqrt{
	\alpha_b t_0
}
$
and the characteristic wave propagation distance
$
\ell_w
=
c_b t_0
$
satisfy
$
\ell_d
\sim
\ell_w.
$
Thus, the diffusion and wave fields interrogate comparable portions of the domain over the selected observation interval while preserving their fundamentally different dynamics.

\begin{figure}[t]
	\includegraphics[width=0.5\textwidth]{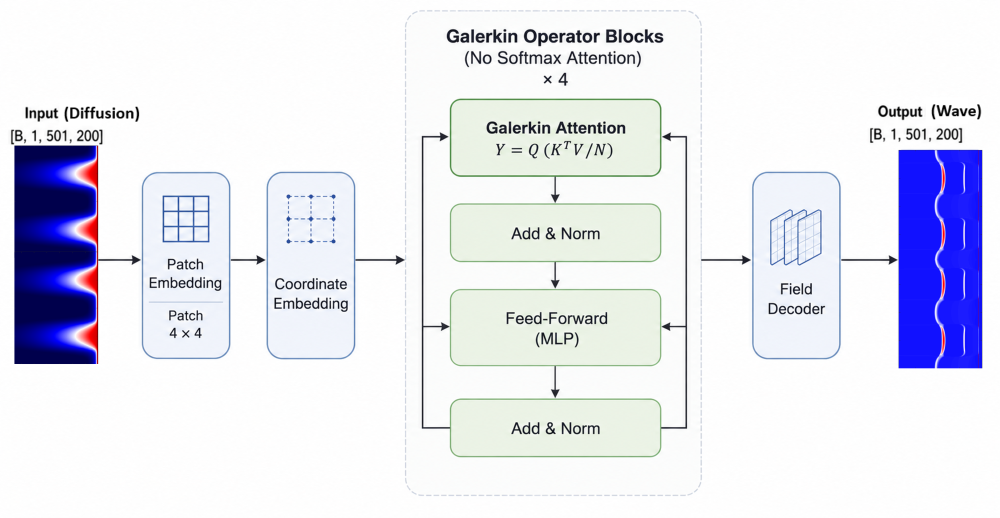}
	\caption{Galerkin Neural Operator (GNO) for cross-physics operator learning.
		Given an input diffusion field
		$T(x,t)\in\mathbb{R}^{501\times200}$,
		GNO predicts the corresponding wave field
		$p(x,t)\in\mathbb{R}^{501\times200}$
		through a sequence of Galerkin operator blocks.
		The input field is first partitioned into $4\times4$ patches and mapped to
		latent tokens, which are augmented with coordinate embeddings to retain
		spatiotemporal positional information.
		Four Galerkin operator blocks then model global interactions among the latent
		field representations.
		Unlike conventional softmax attention, the Galerkin attention evaluates
		$Y=Q(K^{\mathsf T}V/N)$ directly, providing an operator-oriented global mixing
		mechanism without softmax normalization.
		Each block combines Galerkin attention with residual normalization and a
		feed-forward network.
		Finally, a field decoder reconstructs the full-resolution wave field from the
		processed latent representation.
		This architecture provides a global, coordinate-aware alternative to
		convolutional and spectral operator representations for learning the
		diffusion-to-wave transformation.}
	\label{fig7}
\end{figure}

\subsection{Preparation for the training datasets}
Although the PDEs are solved over the complete two-dimensional domain, the learning problem is deliberately restricted to boundary-accessible measurements. We define the observation operator
$
\mathcal{O}_{\Gamma}
:
f(x,y,t)
\mapsto
f(x,0,t),
$
which extracts the field along the front surface $\Gamma_s$.
The diffusion and wave observations are therefore
$
u_{\mathrm{D}}(x,t)
=
\mathcal{O}_{\Gamma}
u(x,y,t),
$
and
$
u_{\mathrm{W}}(x,t)
=
\mathcal{O}_{\Gamma}
v(x,y,t).
$
Hence, the neural networks do not receive the full two-dimensional internal fields or the defect map as direct inputs. They are trained only on the surface measurements
$
u_{\mathrm{D}}(x,t)
\longrightarrow
u_{\mathrm{W}}(x,t).
$
This setting intentionally creates a partial-observation problem. Information about the subsurface structure is encoded only indirectly through its effect on the measured surface dynamics, resembling practical sensing scenarios in which the interior of the object is inaccessible.

The PDE solvers use their own internal temporal steps
$\Delta t_d$ and $\Delta t_w$ to satisfy the corresponding numerical stability requirements. The resulting surface fields are subsequently sampled on a common observation grid
$
N_t^{\mathrm{obs}}
\times
N_x^{\mathrm{obs}}
=
501\times200,
$
which is the resolution used by all learning architectures in this work.
For each realization,
$
\mathbf{U}_{\mathrm{D}}
=
\left[
u_{\mathrm{D}}
(x_j,t_i)
\right]
\in
\mathbb{R}^{501\times200},
$
and
$
\mathbf{U}_{\mathrm{W}}
=
\left[
u_{\mathrm{W}}
(x_j,t_i)
\right]
\in
\mathbb{R}^{501\times200}.
$
Using a common observation grid removes numerical-grid differences between the two solvers and ensures that the neural networks learn a cross-physics transformation rather than a trivial grid interpolation.

To prevent the learned mapping from being restricted to a small family of deterministic configurations, the latent scene parameters are randomized independently for each realization. The randomized variables include the inclusion center, depth, aspect ratio, orientation, material contrast, excitation position, excitation width, amplitude, and selected dimensionless physical parameters. We denote the complete latent parameter vector by
$
\boldsymbol{\theta}^{(n)}
=
\left[
\boldsymbol{\theta}_d^{(n)},
\boldsymbol{\theta}_s^{(n)},
\boldsymbol{\theta}_p^{(n)}
\right],
$
where
$\boldsymbol{\theta}_d$ contains defect parameters,
$\boldsymbol{\theta}_s$ contains source parameters, and
$\boldsymbol{\theta}_p$ contains physical coefficients.
For a realization $\boldsymbol{\theta}^{(n)}$, the paired fields are generated as
$
\mathbf{U}_{\mathrm{D}}^{(n)}
=
\mathcal{O}_{\Gamma}
\mathcal{S}_{d}
\left(
\boldsymbol{\theta}^{(n)}
\right),
$
and
$
\mathbf{U}_{\mathrm{W}}^{(n)}
=
\mathcal{O}_{\Gamma}
\mathcal{S}_{w}
\left(
\boldsymbol{\theta}^{(n)}
\right).
$
The essential pairing condition is therefore
each paired sample $(\mathbf{U}_{\mathrm{D}},\mathbf{U}_{\mathrm{W}})$ is generated under the same latent structure, matched source configuration, and matched dimensionless scale.
A total of
$
N_s=1000
$
paired realizations are generated. The data are randomly partitioned into
$
800/100/100
$
samples for training, validation, and testing, respectively. The split is performed at the realization level such that the diffusion and wave fields generated from the same latent configuration always remain in the same subset.

To avoid information leakage, normalization statistics are computed exclusively from the training subset. Diffusion and wave fields are standardized independently,
$
\widetilde{\mathbf{U}}_{\mathrm{D}}
=
(\mathbf{U}_{\mathrm{D}}
-
\mu_{\mathrm{D}})
/
\sigma_{\mathrm{D}},
$
and
$
\widetilde{\mathbf{U}}_{\mathrm{W}}
=
(\mathbf{U}_{\mathrm{W}}
-
\mu_{\mathrm{W}})
/
\sigma_{\mathrm{W}}.
$
The same training-set statistics are subsequently applied to the validation and test subsets.
The resulting learning task is therefore
$
\widetilde{\mathbf{U}}_{\mathrm{D}}
\in
\mathbb{R}^{501\times200}
\xrightarrow{\;\mathcal{G}_{\theta}\;}
\widetilde{\mathbf{U}}_{\mathrm{W}}
\in
\mathbb{R}^{501\times200}.
$
This dataset design separates the cross-physics learning problem from direct access to the latent geometry. The networks are never provided with the defect map or internal material coefficients; instead, they must infer the wave-domain response exclusively from the diffusion-domain surface dynamics. Consequently, successful prediction requires learning the nonlocal relationship between two distinct evolution operators rather than memorizing an explicit geometric representation.
To improve the robustness and generalization ability of the proposed cross-physics mapping framework, the training dataset was generated under diverse simulation conditions. As summarized in Table~\ref{tab:dataset}, the thermal diffusivity, specimen thickness, and excitation pattern were systematically varied, while the observation duration and material properties were kept consistent. This design allows the learned operator to capture invariant relationships across a broad range of physical scales and excitation mechanisms.

\begin{table}[t] \centering \caption{Simulation configurations used for training, validation, and testing of the cross-physics mapping models. The dataset covers different thermal diffusivities, sample thicknesses, and excitation patterns to improve the diversity and generalization capability of the learned operators.} \label{tab:dataset} \begin{tabular}{lcc} \toprule Parameter & Range / Type & Values \\ \midrule Sample thickness & Variable & 5, 7, 10 mm \\ Excitation source & Variable & Plane wave, Gaussian,\\ & & Asymmetric Gaussian,\\ & & Local Gaussian \\ Thermal diffusivity $\alpha$ & Variable & $10^{-7}$--$5\times10^{-5}\ \mathrm{m^2/s}$ \\ Observation time $\tau$ & Fixed & 10 s \\ Wave velocity $c$ & Derived & $10^{-4}$--$2.2\times10^{-3}\ \mathrm{m/s}$ \\ Scaling parameter $E$ & Derived & $7.4\times10^{-6}$--$3.7\times10^{-3}$ \\ Density $\rho$ & Fixed & $1000\ \mathrm{kg/m^3}$ \\ Poisson ratio $\nu$ & Fixed & 0.3 \\ Thermal frames & Fixed & 100 \\ Wave frames & Fixed & 100 \\ \bottomrule \end{tabular} \end{table}

\begin{figure}[t]
	\includegraphics[width=0.5\textwidth]{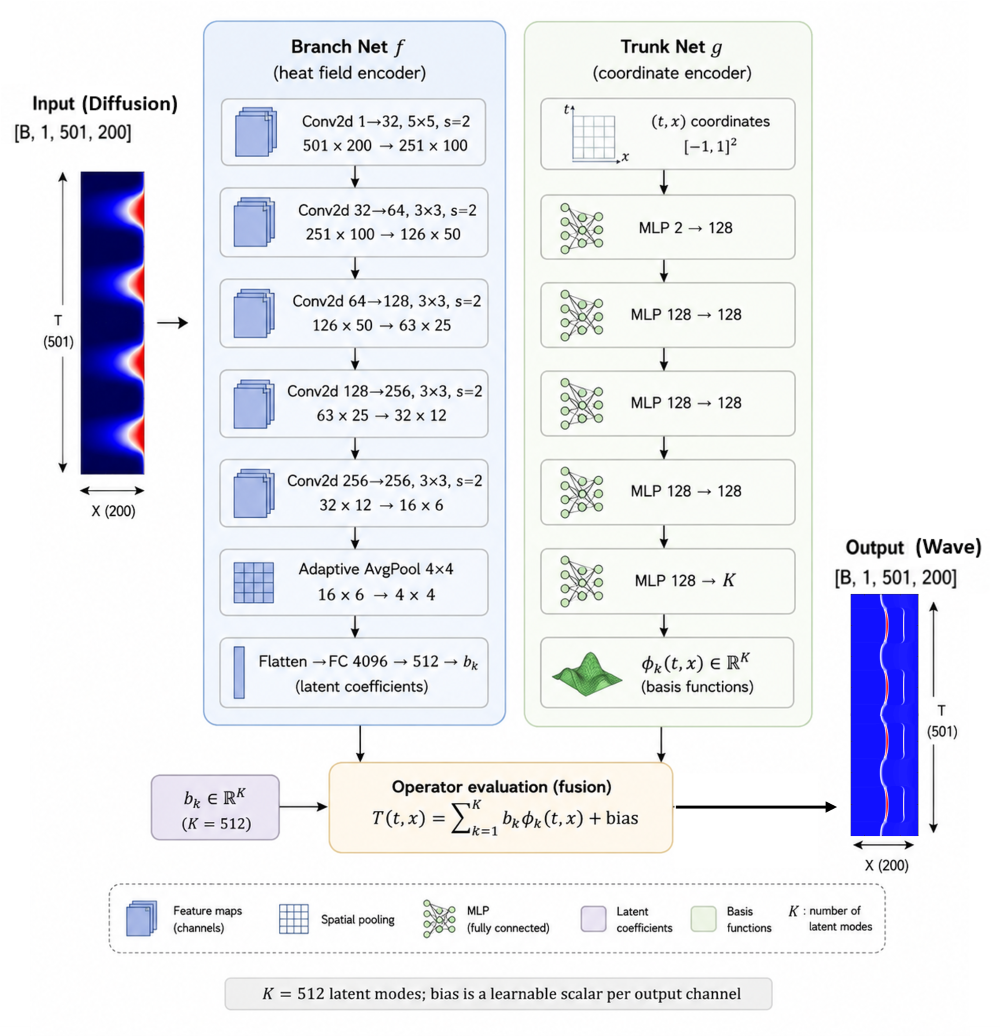}
	\caption{DeepONet for diffusion-to-wave cross-physics operator learning.
		DeepONet represents the mapping from the input diffusion field
		$T(x,t)\in\mathbb{R}^{501\times200}$
		to the corresponding wave field
		$p(x,t)\in\mathbb{R}^{501\times200}$
		through separate branch and trunk networks.
		The branch network encodes the complete input field using a hierarchy of
		convolutional layers and adaptive spatial pooling, producing
		$K=512$ latent coefficients $\{b_k\}_{k=1}^{K}$.
		In parallel, the trunk network maps each normalized spatiotemporal coordinate
		$(t,x)\in[-1,1]^2$ through a multilayer perceptron to
		$K$ coordinate-dependent basis functions
		$\{\phi_k(t,x)\}_{k=1}^{K}$.
		The output field is evaluated pointwise through the operator expansion
		$p(t,x)=\sum_{k=1}^{K} b_k\phi_k(t,x)+b_0$,
		where $b_0$ is a learnable bias.
		This branch--trunk factorization separates input-field encoding from
		coordinate-dependent field evaluation, providing a continuous
		function-to-function representation of the diffusion-to-wave operator.}
	\label{fig8}
\end{figure}

\section{Methodology}

In this work, to achieve a comprehensive evaluation, one image-based learning backbone (ResUNet) and six commonly-used neural operators were employed for cross-physics learning, including U-Shaped Neural Operator (U-NO), Wavelet Neural Operator (WNO), Latent Neural Operator (LNO), Galerkin Neural Operator (GNO), DeepONet, and Fourier Neural Operator (FNO). We made some adjustments so that the neural operators could adapt to the cross-physics mapping.

\subsection{Cross-Physics Mapping Based on ResUNet}
\label{sec:resunet}
We first consider a residual U-Net (ResUNet)~\cite{zhangieee2018} as a convolutional baseline for learning the cross-physics transformation from the diffusion field to the wave field. Unlike image-to-image translation between observations governed by the same physical process, the present task requires the network to approximate a mapping between two fields with distinct governing dynamics. Let
$
\mathbf{u}_{\mathrm{D}}
\in
\mathbb{R}^{1\times N_t\times N_x}
$
denote the measured diffusion field and
$
\mathbf{u}_{\mathrm{W}}
\in
\mathbb{R}^{1\times N_t\times N_x}
$
the corresponding wave field, where $N_t=501$ and $N_x=200$ in our implementation. The ResUNet learns a parameterized mapping
$
\mathcal{G}_{\theta}:
\mathbf{u}_{\mathrm{D}}
\mapsto
\widehat{\mathbf{u}}_{\mathrm{W}},
\widehat{\mathbf{u}}_{\mathrm{W}}
=
\mathcal{G}_{\theta}
\left(
\mathbf{u}_{\mathrm{D}}
\right),
$
where $\theta$ denotes the trainable network parameters. The objective is therefore to directly approximate the cross-physics input--output operator from paired diffusion--wave observations.

\paragraph{Network architecture.}
The adopted ResUNet follows a symmetric encoder--decoder architecture with four resolution levels and a residual bottleneck. The encoder progressively extracts multiscale representations of the diffusion field using feature widths
$
32 \rightarrow 64 \rightarrow 128 \rightarrow 256,
$
followed by a $512$-channel bottleneck. Each encoder stage consists of a residual block followed by $2\times2$ max pooling with stride $2$. The resulting hierarchy enables the network to combine local field variations with increasingly large receptive fields, which is important for capturing nonlocal dependencies between diffusive and propagating responses.
For an input feature tensor $\mathbf{z}$, each residual block is defined as
$
\mathcal{R}(\mathbf{z})
=
\phi
\left[
\mathcal{B}_2
\left(
\mathcal{C}_2
\left(
\phi
\left[
\mathcal{B}_1
\left(
\mathcal{C}_1(\mathbf{z})
\right)
\right]
\right)
\right)
+
\mathcal{S}(\mathbf{z})
\right],
$
where $\mathcal{C}_1$ and $\mathcal{C}_2$ denote $3\times3$ convolutions with unit padding, $\mathcal{B}$ denotes batch normalization, and $\phi(\cdot)$ is the GELU activation. The shortcut operator $\mathcal{S}$ is the identity when the input and output channel dimensions are identical; otherwise, a $1\times1$ convolution projects the input onto the required channel dimension. The residual formulation facilitates gradient propagation through the deep encoder--decoder network while allowing each stage to learn corrections relative to its input representation.
The decoder mirrors the encoder and progressively reconstructs the wave field with channel dimensions
$
512 \rightarrow 256 \rightarrow 128
\rightarrow 64 \rightarrow 32.
$
At each decoder level, the lower-resolution representation is first bilinearly interpolated to the spatial--temporal resolution of the corresponding encoder feature map. A $1\times1$ convolution then reduces its channel dimension, after which the decoder representation is concatenated with the encoder feature through a U-Net skip connection:
$
\mathbf{z}^{d}_{\ell}
=
\mathcal{R}_{\ell}^{d}
\left(
\operatorname{Concat}
\left[
\mathcal{P}_{\ell}
\left(
\operatorname{Interp}
(\mathbf{z}^{d}_{\ell+1})
\right),
\mathbf{z}^{e}_{\ell}
\right]
\right),
\label{eq:decoder}
$
where $\mathbf{z}^{e}_{\ell}$ and $\mathbf{z}^{d}_{\ell}$ denote the encoder and decoder features at level $\ell$, respectively, and $\mathcal{P}_{\ell}$ denotes the $1\times1$ channel projection. These skip connections preserve fine-scale temporal and spatial information that may otherwise be lost during repeated downsampling.
The use of explicit interpolation is particularly important for the present data because the temporal dimension $N_t=501$ is not exactly divisible by $2^4$. Instead of assuming perfectly symmetric feature-map dimensions, each decoder feature is resized directly to the size of its corresponding encoder feature. This guarantees consistent skip concatenation and allows the final prediction to retain the original $501\times200$ resolution. A final $1\times1$ convolution maps the $32$ decoder channels to a single output channel,
$
\widehat{\mathbf{u}}_{\mathrm{W}}
=
\mathcal{C}_{1\times1}
\left(
\mathbf{z}^{d}_{1}
\right),
$
yielding a wave prediction with the same dimensions as the input diffusion field.

\begin{figure}[t]
	\includegraphics[width=0.5\textwidth]{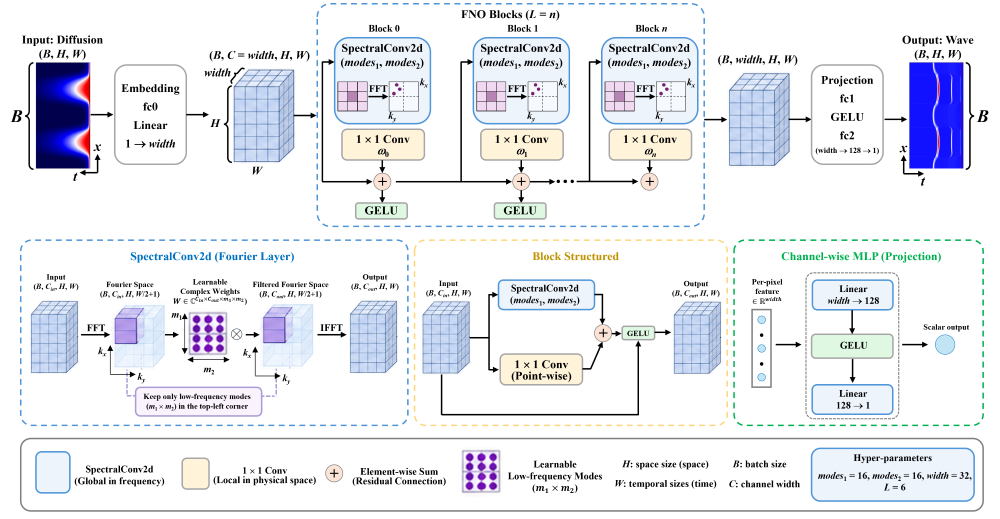}
	\caption{Fourier Neural Operator (FNO) for diffusion-to-wave cross-physics mapping.
		The input diffusion field is first lifted from a scalar field to a
		$\mathrm{width}$-dimensional feature representation and subsequently processed
		by $L=6$ Fourier operator blocks.
		Each block combines a global spectral convolution with a local pointwise
		$1\times1$ convolution and a residual connection, followed by a GELU
		nonlinearity.
		In the spectral branch, a two-dimensional fast Fourier transform (FFT) maps
		the spatiotemporal features to Fourier space, where only the lowest
		$m_1\times m_2$ modes are retained and multiplied by learnable complex-valued
		weights; an inverse FFT then returns the transformed features to the physical
		domain.
		The resulting spectral and pointwise representations are combined to capture
		complementary nonlocal and local interactions.
		Finally, a channel-wise MLP projects the latent representation back to the
		scalar wave field at the original resolution.
		The experiments use $m_1=m_2=16$, $\mathrm{width}=32$, and $L=6$.
		The lower panels detail the spectral convolution, individual operator block,
		and output projection, respectively.}
	\label{fig9}
\end{figure}

\paragraph{Normalization and training objective.}
The diffusion and wave fields are standardized independently using statistics computed exclusively from the training set,
$
\widetilde{\mathbf{u}}_{\mathrm{D}}
=
(\mathbf{u}_{\mathrm{D}}-\mu_{\mathrm{D}})
/
\sigma_{\mathrm{D}},
\widetilde{\mathbf{u}}_{\mathrm{W}}
=
(\mathbf{u}_{\mathrm{W}}-\mu_{\mathrm{W}})
/
\sigma_{\mathrm{W}}.
$
This prevents information leakage from the validation and test sets while reducing the scale disparity between the two physical fields.
The network is trained using a composite objective consisting of the mean-squared error (MSE) and relative $\ell_2$ error,
\begin{equation}
	\begin{split}
	\mathcal{L}_{\mathrm{ResUNet}}
	&=
	\mathcal{L}_{\mathrm{MSE}}
	+
	\lambda_{\mathrm{rel}}
	\mathcal{L}_{\mathrm{rel}},
	\quad
	\mathcal{L}_{\mathrm{MSE}}
	=
	\frac{1}{N}
	\left\|
	\widehat{\mathbf{u}}_{\mathrm{W}}
	-
	\mathbf{u}_{\mathrm{W}}
	\right\|_2^2, 
	\\
	\mathcal{L}_{\mathrm{rel}}
	&=
	\frac{
		\left\|
		\widehat{\mathbf{u}}_{\mathrm{W}}
		-
		\mathbf{u}_{\mathrm{W}}
		\right\|_2
	}{
		\left\|
		\mathbf{u}_{\mathrm{W}}
		\right\|_2+\epsilon
	}.
	\end{split}
	\label{eq:resunet_loss}
\end{equation}

We set $\lambda_{\mathrm{rel}}=0.1$ and $\epsilon=10^{-8}$. The MSE term penalizes pointwise reconstruction errors, whereas the relative $\ell_2$ term normalizes the discrepancy by the magnitude of the target field and therefore provides a scale-aware measure of global field reconstruction.
Optimization is performed using AdamW with an initial learning rate of $10^{-3}$ and a weight decay of $10^{-5}$. A cosine-annealing schedule gradually decreases the learning rate during training, and the gradient norm is clipped at $1.0$ for numerical stability. The model yielding the lowest validation relative $\ell_2$ error is retained for final evaluation.
From a cross-physics perspective, ResUNet provides a strong local convolutional baseline: the encoder captures hierarchical diffusion features, the bottleneck aggregates large-scale contextual information, and the decoder reconstructs the target wave dynamics while recovering fine-scale structures through skip connections. However, its mapping is fundamentally constructed from finite-receptive-field convolutional operations. We therefore use ResUNet as a reference architecture against which the subsequent neural-operator models are evaluated, particularly with respect to their ability to learn nonlocal and resolution-independent mappings between distinct physical systems.

\subsection{Cross-Physics Mapping Based on U-Shaped Neural Operator (U-NO)}
\label{sec:uno}
To move beyond the finite receptive field of convolutional architectures, we further consider a U-shaped Neural Operator (U-NO)~\cite{Rahman2022} for learning the diffusion-to-wave cross-physics transformation. In contrast to ResUNet, which primarily propagates information through local convolutional kernels, U-NO performs global spectral mixing at multiple resolutions. This provides a natural mechanism for representing the nonlocal dependencies required to transform a diffusion field into a propagating wave field.
Given a normalized diffusion field
$
\mathbf{u}_{\mathrm{D}}
\in
\mathbb{R}^{1\times N_t\times N_x},
$
the U-NO learns the operator
$
\mathcal{G}_{\theta}^{\mathrm{UNO}}:
\mathbf{u}_{\mathrm{D}}
\mapsto
\widehat{\mathbf{u}}_{\mathrm{W}},
\widehat{\mathbf{u}}_{\mathrm{W}}
=
\mathcal{G}_{\theta}^{\mathrm{UNO}}
\left(
\mathbf{u}_{\mathrm{D}}
\right),
$
where $N_t=501$ and $N_x=200$ in our experiments. Rather than learning this transformation exclusively through local feature extraction, U-NO combines spectral integral operators, pointwise mappings, and multiresolution skip connections.

\paragraph{Coordinate lifting.}
Neural operators benefit from explicit positional information because the operator acts on functions defined over a physical domain. We therefore augment the diffusion field with normalized temporal and spatial coordinates,
$
\mathbf{g}(t,x)
=
\left[
\widetilde{t},
\widetilde{x}
\right],
\widetilde{t},\widetilde{x}\in[0,1],
$
and construct the lifted input as
$
\mathbf{z}_0
=
\mathcal{P}
\left(
\operatorname{Concat}
[
\mathbf{u}_{\mathrm{D}},
\widetilde{t},
\widetilde{x}
]
\right),
$
where $\mathcal{P}$ is a $1\times1$ convolution that lifts the three-channel input representation to a latent width of $32$. The coordinate channels allow the network to distinguish identical field values occurring at different temporal or spatial locations.

\paragraph{Spectral operator block.}
The fundamental component of U-NO is a two-dimensional neural-operator block operating jointly along the temporal and spatial dimensions. For a feature representation $\mathbf{z}_{\ell}$, the block is written as
$
\mathbf{z}_{\ell+1}
=
\phi
\left[
\mathcal{N}_{\ell}
\left(
\mathcal{K}_{\ell}
(\mathbf{z}_{\ell})
+
\mathcal{W}_{\ell}
(\mathbf{z}_{\ell})
\right)
\right],
$
where $\mathcal{K}_{\ell}$ denotes the spectral operator, $\mathcal{W}_{\ell}$ is a learned pointwise $1\times1$ convolution, $\mathcal{N}_{\ell}$ denotes group normalization, and $\phi(\cdot)$ is the GELU activation. The spectral and local branches are complementary: the former captures nonlocal interactions over the entire field, whereas the latter preserves local channel-wise transformations.
The spectral branch first transforms the latent representation into the Fourier domain,
$
\widehat{\mathbf{z}}_{\ell}
=
\mathcal{F}
\left(
\mathbf{z}_{\ell}
\right),
$
where $\mathcal{F}$ denotes the two-dimensional Fourier transform over $(t,x)$. A learned complex-valued spectral kernel is then applied to a truncated set of Fourier modes,
$
\widehat{\mathbf{v}}_{\ell}(k_t,k_x)
=
\mathbf{R}_{\ell}(k_t,k_x)
\widehat{\mathbf{z}}_{\ell}(k_t,k_x),
(k_t,k_x)\in\mathcal{M}_{\ell},
$
where $\mathbf{R}_{\ell}$ is a trainable complex-valued tensor and $\mathcal{M}_{\ell}$ denotes the retained spectral modes. The spectral output is obtained through the inverse transform,
$
\mathcal{K}_{\ell}
(\mathbf{z}_{\ell})
=
\mathcal{F}^{-1}
\left(
\widehat{\mathbf{v}}_{\ell}
\right).
$
In our implementation, separate learnable weights are used for positive and negative temporal frequencies, while the real-valued FFT representation is retained along the spatial dimension. This construction enables direct global communication between distant temporal--spatial locations within a single operator layer.

\paragraph{Multiscale U-shaped operator architecture.}
The U-NO organizes these operator blocks into an encoder--bottleneck--decoder hierarchy. Starting from the lifted $32$-channel representation, the encoder uses three operator levels with channel widths
$
32 \rightarrow 64 \rightarrow 128,
$
followed by a $256$-channel bottleneck. Each encoder operator is evaluated before resolution reduction, and its output is retained as a skip feature. Downsampling is performed using $2\times2$ average pooling with stride $2$.
Importantly, the number of retained Fourier modes is progressively reduced with resolution. The temporal--spatial mode pairs used in the encoder are
$
(M_t,M_x)
=
(64,40),
(32,24),
(16,12),
$
while the bottleneck retains
$
(M_t,M_x)=(8,6).
$
The number of modes is automatically clipped when the feature-map resolution becomes smaller than the requested spectral bandwidth. This multiscale spectral hierarchy allocates a larger Fourier bandwidth to high-resolution representations while allowing coarse levels to model large-scale interactions using a compact set of modes.
The decoder reverses this hierarchy,
$
256 \rightarrow 128
\rightarrow 64
\rightarrow 32.
$
At decoder level $\ell$, the coarse representation is bilinearly interpolated to exactly match the corresponding encoder resolution and subsequently reduced in channel dimension through a $1\times1$ projection. The resulting representation is concatenated with the encoder feature,
$
\widetilde{\mathbf{z}}_{\ell}^{d}
=
\operatorname{Concat}
\left[
\mathcal{P}_{\ell}^{d}
\left(
\operatorname{Interp}
(
\mathbf{z}_{\ell+1}^{d}
)
\right),
\mathbf{z}_{\ell}^{e}
\right],
$
and processed by another neural-operator block,
$
\mathbf{z}_{\ell}^{d}
=
\mathcal{O}_{\ell}^{d}
\left(
\widetilde{\mathbf{z}}_{\ell}^{d}
\right).
$
The U-shaped skip pathways therefore transfer high-resolution features directly to the decoder, while the spectral operator blocks simultaneously model long-range dependencies at each scale.
Because $N_t=501$ is odd, repeated factor-of-two rescaling does not produce perfectly symmetric encoder and decoder dimensions. We use average pooling with \texttt{ceil\_mode=True} during encoding and explicitly interpolate each decoder feature to the corresponding skip-feature size. This avoids artificial cropping or padding and guarantees reconstruction at the original $501\times200$ resolution.

\paragraph{Local refinement and output projection.}
After the final decoder stage, a local residual refinement module is applied,
$
\mathbf{z}_{r}
=
\mathbf{z}_{d}
+
\mathcal{R}_{\mathrm{local}}
(\mathbf{z}_{d}),
$
where $\mathcal{R}_{\mathrm{local}}$ contains two $3\times3$ convolutional layers with GELU activations. This refinement complements the global spectral representation by correcting local reconstruction details.
The refined $32$-channel representation is finally projected through two pointwise layers,
$
\widehat{\mathbf{u}}_{\mathrm{W}}
=
\mathcal{Q}_2
\left[
\phi
\left(
\mathcal{Q}_1
(\mathbf{z}_{r})
\right)
\right],
$
where $\mathcal{Q}_1$ maps $32$ channels to $128$ latent channels and $\mathcal{Q}_2$ maps the representation to the single-channel wave field.

\paragraph{Training objective.}
For consistency across architectures, U-NO is trained using the same composite objective as the convolutional baseline,
\begin{equation}
	\begin{split}
	\mathcal{L}_{\mathrm{UNO}}
	&=
	\mathcal{L}_{\mathrm{MSE}}
	+
	\lambda_{\mathrm{rel}}
	\mathcal{L}_{\mathrm{rel}},
	\qquad
	\lambda_{\mathrm{rel}}=0.1,
	\quad
	\\
	\mathcal{L}_{\mathrm{rel}}
	&=
	\frac{
		\left\|
		\widehat{\mathbf{u}}_{\mathrm{W}}
		-
		\mathbf{u}_{\mathrm{W}}
		\right\|_2
	}{
		\left\|
		\mathbf{u}_{\mathrm{W}}
		\right\|_2+\epsilon
	},
	\qquad
	\epsilon=10^{-8}.
	\end{split}
	\label{eq:uno_loss}
\end{equation}

The diffusion and wave fields are independently standardized using statistics computed only from the training data. We use AdamW with an initial learning rate of $10^{-3}$, weight decay of $10^{-5}$, cosine learning-rate annealing, and gradient clipping with a maximum norm of $1.0$. The checkpoint with the lowest validation relative $\ell_2$ error is used for final testing.
The principal distinction between U-NO and ResUNet lies in how information is propagated across the field. ResUNet builds nonlocal context indirectly through hierarchical local convolutions, whereas U-NO performs explicit global mixing in the Fourier domain at every resolution. Its U-shaped multiscale organization further combines global operator learning with high-resolution skip features. U-NO therefore provides a particularly relevant neural-operator baseline for evaluating whether explicit nonlocal representations improve mappings between physical systems governed by qualitatively different dynamics.

\subsection{Cross-Physics Mapping Based on Wavelet Neural Operator (WNO)}
\label{sec:wno}
To further investigate whether localized multiscale representations are beneficial for cross-physics learning, we employ a Wavelet Neural Operator (WNO)~\cite{Tripura2023} to learn the transformation from the diffusion field to the wave field. Unlike Fourier-based neural operators, which represent the operator through globally supported Fourier modes, WNO performs operator learning in a wavelet domain that is localized in both scale and position. This property is particularly relevant to the present diffusion-to-wave mapping, where slowly varying diffusive components coexist with localized and rapidly varying wave-like structures.
Given a normalized diffusion field
$
\mathbf{u}_{\mathrm{D}}
\in
\mathbb{R}^{1\times N_t\times N_x},
$
the WNO learns the nonlinear operator
$
\mathcal{G}_{\theta}^{\mathrm{WNO}}:
\mathbf{u}_{\mathrm{D}}
\mapsto
\widehat{\mathbf{u}}_{\mathrm{W}},
\widehat{\mathbf{u}}_{\mathrm{W}}
=
\mathcal{G}_{\theta}^{\mathrm{WNO}}
\left(
\mathbf{u}_{\mathrm{D}}
\right),
$
where $N_t=501$ and $N_x=200$ in our experiments. The central idea is to replace global Fourier-domain mixing with learnable multiresolution transformations of wavelet coefficients.

\paragraph{Coordinate lifting.}
As in the other neural-operator architectures, the input field is augmented with its normalized temporal and spatial coordinates,
$
\mathbf{g}(t,x)
=
\left[
\widetilde{t},
\widetilde{x}
\right],
\widetilde{t},\widetilde{x}\in[0,1].
$
The resulting three-channel representation is lifted to a latent feature space according to
$
\mathbf{z}_0
=
\mathcal{P}
\left(
\operatorname{Concat}
[
\mathbf{u}_{\mathrm{D}},
\widetilde{t},
\widetilde{x}
]
\right),
$
where $\mathcal{P}$ denotes a $1\times1$ convolution. In our implementation, the latent width is set to $48$. Explicit coordinate information allows the learned operator to distinguish identical field amplitudes occurring at different temporal--spatial locations.

\paragraph{Wavelet decomposition.}
The principal component of WNO is a multilevel two-dimensional Haar wavelet transform. For a latent feature field $\mathbf{z}$, one level of the discrete wavelet transform (DWT) decomposes the representation into four subbands,
$
\mathcal{W}(\mathbf{z})
=
\left(
\mathbf{z}_{LL},
\mathbf{z}_{LH},
\mathbf{z}_{HL},
\mathbf{z}_{HH}
\right),
$
where $LL$ denotes the low-frequency approximation coefficients, while $LH$, $HL$, and $HH$ contain localized detail information associated with different temporal--spatial variations.
For the orthonormal Haar transform used here, the four $2\times2$ analysis kernels can be expressed as
\begin{equation}
	\begin{split}
	\mathbf{H}_{LL}
	&=
	\frac{1}{2}
	\begin{bmatrix}
		1 & 1\\
		1 & 1
	\end{bmatrix},
	\quad
	\mathbf{H}_{LH}
	=
	\frac{1}{2}
	\begin{bmatrix}
		-1 & -1\\
		1 &  1
	\end{bmatrix},
	\\
	\mathbf{H}_{HL}
	&=
	\frac{1}{2}
	\begin{bmatrix}
		-1 & 1\\
		-1 & 1
	\end{bmatrix},
	\quad
	\mathbf{H}_{HH}
	=
	\frac{1}{2}
	\begin{bmatrix}
		1 & -1\\
		-1 &  1
	\end{bmatrix}.
	\end{split}
	\label{eq:haar_filters}
\end{equation}
The transform is implemented using grouped convolutions with stride $2$, such that each latent channel is decomposed independently. Consequently, the representation is simultaneously separated according to spatial--temporal scale and local variation.

\paragraph{Learnable wavelet-domain operator.}
Instead of directly modifying the field in the physical domain, WNO applies independent learnable transformations to the wavelet subbands. At decomposition level $\ell$, we write
$
\widetilde{\mathbf{z}}_{b}^{(\ell)}
=
\mathcal{R}_{b}^{(\ell)}
\left(
\mathbf{z}_{b}^{(\ell)}
\right),
b\in\{LL,LH,HL,HH\},
$
where $\mathcal{R}_{b}^{(\ell)}$ is a trainable channel-mixing operator associated with subband $b$. In our implementation, each $\mathcal{R}_{b}^{(\ell)}$ is parameterized by an independent $1\times1$ convolution.
This decomposition provides a direct mechanism for learning different transformations for coarse and fine field components. In particular, the $LL$ branch represents slowly varying large-scale structures, whereas the detail bands encode localized temporal--spatial variations. Unlike a single global spectral kernel, the wavelet representation therefore retains information about both scale and localization.

\paragraph{Multilevel wavelet operator.}
To construct a hierarchical multiscale representation, the wavelet decomposition is recursively applied to the approximation coefficients. Denoting the approximation at level $\ell$ by $\mathbf{z}_{LL}^{(\ell)}$, the next level is obtained as
$
\left(
\mathbf{z}_{LL}^{(\ell+1)},
\mathbf{z}_{LH}^{(\ell+1)},
\mathbf{z}_{HL}^{(\ell+1)},
\mathbf{z}_{HH}^{(\ell+1)}
\right)
=
\mathcal{W}
\left(
\widetilde{\mathbf{z}}_{LL}^{(\ell)}
\right).
$
We employ two levels of Haar decomposition. The detail coefficients at each scale are retained, while the transformed $LL$ coefficients are recursively propagated to the coarser level.
After wavelet-domain mixing, reconstruction proceeds in reverse order using the inverse discrete wavelet transform (IDWT),
$
\mathbf{z}^{(\ell)}
=
\mathcal{W}^{-1}
\left(
\widetilde{\mathbf{z}}_{LL}^{(\ell)},
\widetilde{\mathbf{z}}_{LH}^{(\ell)},
\widetilde{\mathbf{z}}_{HL}^{(\ell)},
\widetilde{\mathbf{z}}_{HH}^{(\ell)}
\right).
$
This decomposition--transformation--reconstruction procedure defines the wavelet integral operator
$
\mathcal{K}_{\ell}^{\mathrm{W}}
=
\mathcal{W}^{-1}
\circ
\mathcal{R}_{\ell}
\circ
\mathcal{W}.
$
Because the temporal dimension $N_t=501$ is odd, the input cannot be repeatedly divided by two without size mismatch. Before each Haar decomposition, odd dimensions are therefore extended by replication padding. After inverse reconstruction, the feature field is cropped to its pre-decomposition size. This procedure guarantees exact recovery of the original $501\times200$ resolution without requiring the input dimensions to be powers of two.

\paragraph{WNO block.}
Each WNO block combines the multiscale wavelet operator with a local pointwise branch. For the latent representation $\mathbf{z}_{\ell}$, we define
$
\mathbf{h}_{\ell}
=
\phi
\left[
\mathcal{N}_{\ell}
\left(
\mathcal{K}_{\ell}^{\mathrm{W}}
(\mathbf{z}_{\ell})
+
\mathcal{P}_{\ell}
(\mathbf{z}_{\ell})
\right)
\right],
$
where $\mathcal{P}_{\ell}$ is a learned $1\times1$ pointwise convolution, $\mathcal{N}_{\ell}$ denotes group normalization, and $\phi(\cdot)$ is the GELU activation. The wavelet branch captures multiscale interactions, while the pointwise branch preserves direct local channel mixing.
A $3\times3$ convolution is subsequently used for local refinement through a residual connection,
$
\mathbf{z}_{\ell+1}
=
\phi
\left[
\mathbf{h}_{\ell}
+
\mathcal{C}_{3\times3}^{(\ell)}
\left(
\mathbf{h}_{\ell}
\right)
\right].
$
This additional residual refinement compensates for the fact that the wavelet-domain transformations themselves primarily perform channel mixing within individual subbands, while the $3\times3$ convolution explicitly restores local interactions between neighboring temporal--spatial locations.

\paragraph{Overall architecture.}
The complete WNO consists of four consecutive WNO blocks operating at a fixed latent width of $48$,
$
\mathbf{z}_{\ell+1}
=
\mathcal{B}_{\ell}^{\mathrm{WNO}}
\left(
\mathbf{z}_{\ell}
\right),
\ell=0,\ldots,3.
$
Each block independently performs a two-level Haar decomposition, learnable subband mixing, inverse reconstruction, local pointwise mixing, and residual refinement. Unlike U-NO, the network does not construct a separate encoder--decoder hierarchy; multiresolution processing is instead performed internally by the wavelet transform within each operator block.
After the final WNO block, the latent representation is projected back to the target wave field using two pointwise layers,
$
\widehat{\mathbf{u}}_{\mathrm{W}}
=
\mathcal{Q}_2
\left[
\phi
\left(
\mathcal{Q}_1
(\mathbf{z}_{4})
\right)
\right],
$
where $\mathcal{Q}_1$ maps the $48$ latent channels to $128$ channels and $\mathcal{Q}_2$ maps the representation to a single output channel.

\paragraph{Training objective.}
For a controlled comparison with the other architectures, WNO is trained using the same composite objective,
\begin{equation}
	\begin{split}
	\mathcal{L}_{\mathrm{WNO}}
	&=
	\mathcal{L}_{\mathrm{MSE}}
	+
	\lambda_{\mathrm{rel}}
	\mathcal{L}_{\mathrm{rel}},
	\qquad
	\lambda_{\mathrm{rel}}=0.1,
	\quad
	\\
	\mathcal{L}_{\mathrm{rel}}
	&=
	\frac{
		\left\|
		\widehat{\mathbf{u}}_{\mathrm{W}}
		-
		\mathbf{u}_{\mathrm{W}}
		\right\|_2
	}{
		\left\|
		\mathbf{u}_{\mathrm{W}}
		\right\|_2+\epsilon
	},
	\qquad
	\epsilon=10^{-8}.
	\end{split}
	\label{eq:wno_loss}
\end{equation}

The diffusion and wave fields are independently standardized using statistics computed exclusively from the training set. Optimization is performed using AdamW with an initial learning rate of $10^{-3}$ and weight decay of $10^{-5}$, together with cosine learning-rate annealing and gradient clipping at a maximum norm of $1.0$. Early stopping is applied according to the validation relative $\ell_2$ error, with a patience of $20$ epochs, and the model with the lowest validation error is retained for testing.
The key distinction between WNO and the preceding architectures lies in the representation used for operator learning. ResUNet constructs the mapping through hierarchical local convolutions, whereas U-NO performs global mixing through truncated Fourier modes. WNO instead represents the field through localized multiresolution wavelet coefficients and learns separate transformations for approximation and detail components. It therefore provides a complementary neural-operator baseline for assessing whether simultaneous localization in scale and position facilitates learning mappings between diffusive diffusion dynamics and propagating wave dynamics.

\subsection{Cross-Physics Mapping Based on Latent Neural Operator (LNO)}
\label{sec:lno}
To investigate whether the cross-physics transformation can be learned more efficiently in a compressed representation space, we further employ a Latent Neural Operator (LNO)~\cite{Wang2024}. In contrast to neural operators that perform global operator learning directly on the full-resolution physical field, LNO first encodes the diffusion field into a compact latent representation, learns the diffusion-to-wave transformation entirely in this latent space, and subsequently decodes the transformed latent representation into the wave field. The overall mapping can therefore be expressed as
$
\widehat{\mathbf{u}}_{\mathrm{W}}
=
\mathcal{D}_{\mathrm{W}}
\circ
\mathcal{G}_{\theta}^{\mathrm{L}}
\circ
\mathcal{E}_{\mathrm{D}}
\left(
\mathbf{u}_{\mathrm{D}}
\right),
$
where $\mathcal{E}_{\mathrm{D}}$ denotes the diffusion encoder,
$\mathcal{G}_{\theta}^{\mathrm{L}}$ is the latent neural operator, and
$\mathcal{D}_{\mathrm{W}}$ denotes the wave decoder. In our experiments,
$
\mathbf{u}_{\mathrm{D}},
\mathbf{u}_{\mathrm{W}}
\in
\mathbb{R}^{1\times501\times200}.
$
The central hypothesis is that the essential transformation between the two physical systems can be represented by a lower-dimensional operator acting on compressed field features rather than directly on the full-resolution observations.

\paragraph{Coordinate-augmented diffusion encoding.}
The input diffusion field is first augmented with normalized temporal and spatial coordinates,
$
\mathbf{g}(t,x)
=
\left[
\widetilde{t},
\widetilde{x}
\right],
\widetilde{t},\widetilde{x}\in[0,1],
$
such that the encoder receives
$
\mathbf{v}_{0}
=
\operatorname{Concat}
\left[
\mathbf{u}_{\mathrm{D}},
\widetilde{t},
\widetilde{x}
\right]
\in
\mathbb{R}^{3\times N_t\times N_x}.
$
The coordinate channels provide explicit positional information and allow the latent representation to distinguish identical diffusion amplitudes occurring at different temporal--spatial locations.
The diffusion encoder progressively compresses the input field through four convolutional stages with channel dimensions
$
3
\rightarrow
32
\rightarrow
64
\rightarrow
96
\rightarrow
128.
$
The first convolutional block operates at the original resolution, while the subsequent stages are separated by $2\times2$ average pooling with stride $2$ and \texttt{ceil\_mode=True}. The corresponding resolutions are approximately
$
501\times200
\rightarrow
251\times100
\rightarrow
126\times50
\rightarrow
63\times25.
$
The resulting latent diffusion field is therefore
$
\mathbf{z}_{\mathrm{D}}
=
\mathcal{E}_{\mathrm{D}}
(\mathbf{v}_{0})
\in
\mathbb{R}^{128\times63\times25}.
$
Each encoder stage employs a residual convolutional block. For an input feature $\mathbf{z}$, the block is written as
$
\mathcal{R}(\mathbf{z})
=
\mathcal{F}(\mathbf{z})
+
\mathcal{S}(\mathbf{z}),
$
where $\mathcal{F}$ contains two consecutive $3\times3$ convolutions, each followed by group normalization and GELU activation, and $\mathcal{S}$ denotes an identity shortcut or a $1\times1$ channel projection when the input and output dimensions differ. This encoder transforms the original physical field into a compact feature representation while retaining nonlinear local structures relevant to the subsequent operator mapping.

\paragraph{Latent spectral operator.}
The central distinction of LNO is that global spectral mixing is performed exclusively in the compressed latent space. For a latent feature
$\mathbf{z}_{\ell}\in
\mathbb{R}^{C_L\times T_L\times X_L}$,
with $C_L=128$, the spectral branch first computes the two-dimensional Fourier transform,
$
\widehat{\mathbf{z}}_{\ell}
=
\mathcal{F}
\left(
\mathbf{z}_{\ell}
\right).
$
A trainable complex-valued kernel is applied to a truncated set of latent Fourier modes,
$
\widehat{\mathbf{v}}_{\ell}(k_t,k_x)
=
\mathbf{R}_{\ell}(k_t,k_x)
\widehat{\mathbf{z}}_{\ell}(k_t,k_x),
(k_t,k_x)\in\mathcal{M}_{L},
$
where $\mathbf{R}_{\ell}$ denotes the learned spectral kernel and
$\mathcal{M}_{L}$ is the retained latent spectral domain.
In our implementation, the nominal number of retained modes is
$
(M_t,M_x)=(24,12).
$
Separate complex-valued weights are learned for positive and negative temporal frequencies, while the real-valued FFT representation is used along the spatial dimension. The number of active modes is automatically restricted according to the actual latent resolution. The spectral output is reconstructed as
$
\mathcal{K}_{\ell}^{L}
(\mathbf{z}_{\ell})
=
\mathcal{F}^{-1}
\left[
\mathbf{R}_{\ell}
\mathcal{F}
(\mathbf{z}_{\ell})
\right].
$
Performing this operation at approximately $63\times25$ rather than
$501\times200$ substantially reduces the spatial--temporal size on which the global operator acts. More importantly, the spectral kernel operates on learned latent features rather than directly on the raw diffusion field. The encoder can therefore first extract representations relevant to the cross-physics transformation, after which the operator models their nonlocal interactions.

\paragraph{Latent operator block.}
Each latent operator block combines a global spectral branch and a local pointwise branch. Given $\mathbf{z}_{\ell}$, we first compute
$
\mathbf{h}_{\ell}
=
\phi
\left[
\mathcal{N}_{\ell}
\left(
\mathcal{K}_{\ell}^{L}
(\mathbf{z}_{\ell})
+
\mathcal{W}_{\ell}
(\mathbf{z}_{\ell})
\right)
\right],
$
where $\mathcal{W}_{\ell}$ is a learned $1\times1$ convolution,
$\mathcal{N}_{\ell}$ denotes group normalization, and
$\phi(\cdot)$ is the GELU activation. The spectral branch provides global latent-space interactions, whereas the pointwise branch preserves direct local channel mixing.
A $3\times3$ convolution is subsequently applied as a local refinement,
$
\widetilde{\mathbf{h}}_{\ell}
=
\mathbf{h}_{\ell}
+
\mathcal{C}_{3\times3}^{(\ell)}
(\mathbf{h}_{\ell}),
$
followed by an outer residual connection,
$
\mathbf{z}_{\ell+1}
=
\phi
\left(
\widetilde{\mathbf{h}}_{\ell}
+
\mathbf{z}_{\ell}
\right).
$
This combination of spectral mixing, pointwise transformation, local refinement, and residual propagation allows each operator block to jointly model global and local dependencies within the compressed field representation.

\paragraph{Cross-physics transformation in latent space.}
Four latent operator blocks are stacked sequentially,
$
\mathbf{z}_{\ell+1}
=
\mathcal{B}_{\ell}^{L}
(\mathbf{z}_{\ell}),
\ell=0,\ldots,3,
$
with
$
\mathbf{z}_{0}
=
\mathbf{z}_{\mathrm{D}}.
$
After global latent processing, an additional pointwise transformation is introduced to explicitly parameterize the transition from the diffusion latent representation to a wave latent representation:
$
\mathbf{z}_{\mathrm{W}}
=
\mathbf{z}_{4}
+
\mathcal{P}_{\mathrm{cross}}
(\mathbf{z}_{4}),
$
where
$
\mathcal{P}_{\mathrm{cross}}
=
\mathcal{C}_{1\times1}^{(2)}
\circ
\phi
\circ
\mathcal{C}_{1\times1}^{(1)}.
$
Both pointwise layers preserve the $128$-channel latent width. This additional residual projection provides an explicit learnable transformation between the latent representations associated with the two physical modalities.
The complete latent transformation can therefore be summarized as
$
\mathbf{z}_{\mathrm{D}}
\xrightarrow{\;
	\mathcal{G}_{\theta}^{L}
	\;}
\mathbf{z}_{\mathrm{W}},
$
rather than requiring the neural operator to directly approximate the full-resolution mapping
$\mathbf{u}_{\mathrm{D}}\rightarrow\mathbf{u}_{\mathrm{W}}$.

\paragraph{Wave decoder.}
The transformed latent representation is decoded progressively back to the physical wave domain. Starting from approximately $63\times25$, bilinear interpolation reconstructs the field through
$
63\times25
\rightarrow
126\times50
\rightarrow
251\times100
\rightarrow
501\times200,
$
while the feature channels are reduced according to
$
128
\rightarrow
96
\rightarrow
64
\rightarrow
32.
$
Each interpolation stage is followed by the same residual convolutional block used in the encoder. Importantly, the decoder does not employ encoder--decoder skip connections. Consequently, the wave reconstruction must be generated from the transformed latent representation rather than directly copying high-resolution diffusion features into the output pathway.
At the original resolution, two additional $3\times3$ convolutions refine the reconstructed field,
$
32
\rightarrow
32
\rightarrow
16,
$
and a final $1\times1$ projection produces the single-channel wave prediction,
$
\widehat{\mathbf{u}}_{\mathrm{W}}
=
\mathcal{D}_{\mathrm{W}}
\left(
\mathbf{z}_{\mathrm{W}}
\right)
\in
\mathbb{R}^{1\times501\times200}.
$
Explicit interpolation to the target resolutions avoids dimensional inconsistencies introduced by the odd temporal dimension $N_t=501$.

\paragraph{Training objective.}
For consistency with the preceding architectures, the diffusion and wave fields are independently standardized using statistics computed exclusively from the training set,
$
\widetilde{\mathbf{u}}_{\mathrm{D}}
=
(\mathbf{u}_{\mathrm{D}}-\mu_{\mathrm{D}})
/
\sigma_{\mathrm{D}},
\widetilde{\mathbf{u}}_{\mathrm{W}}
=
(\mathbf{u}_{\mathrm{W}}-\mu_{\mathrm{W}})
/
\sigma_{\mathrm{W}}.
$
LNO is optimized using the same composite reconstruction objective,
\begin{equation}
	\begin{split}
	\mathcal{L}_{\mathrm{LNO}}
	&=
	\mathcal{L}_{\mathrm{MSE}}
	+
	\lambda_{\mathrm{rel}}
	\mathcal{L}_{\mathrm{rel}},
	\qquad
	\lambda_{\mathrm{rel}}=0.1,
	\quad
	\\
	\mathcal{L}_{\mathrm{rel}}
	&=
	\frac{
		\left\|
		\widehat{\mathbf{u}}_{\mathrm{W}}
		-
		\mathbf{u}_{\mathrm{W}}
		\right\|_2
	}{
		\left\|
		\mathbf{u}_{\mathrm{W}}
		\right\|_2+\epsilon
	},
	\qquad
	\epsilon=10^{-8}.
	\end{split}
	\label{eq:lno_loss}
\end{equation}

We use AdamW with an initial learning rate of $10^{-3}$ and weight decay of $10^{-5}$. Cosine learning-rate annealing is applied over training, and the gradient norm is clipped at $1.0$. The model corresponding to the minimum validation relative $\ell_2$ error is retained for final evaluation.
The defining characteristic of LNO is therefore not merely the use of Fourier layers, but the location at which operator learning is performed. U-NO applies spectral operators directly within a multiresolution physical-field hierarchy, whereas LNO first separates representation learning from operator learning:
$
\mathbf{u}_{\mathrm{D}}
\xrightarrow{\mathcal{E}_{\mathrm{D}}}
\mathbf{z}_{\mathrm{D}}
\xrightarrow{\mathcal{G}_{\theta}^{L}}
\mathbf{z}_{\mathrm{W}}
\xrightarrow{\mathcal{D}_{\mathrm{W}}}
\widehat{\mathbf{u}}_{\mathrm{W}}.
$
This architecture tests whether the diffusion-to-wave transformation admits a compact latent representation in which the essential cross-physics relation can be modeled through a global neural operator. It therefore provides a complementary perspective to ResUNet, U-NO, and WNO: rather than asking only which representation is most effective for operator learning, LNO additionally examines whether the cross-physics operator itself can be learned after substantial dimensional compression.

\subsection{Cross-Physics Mapping Based on Galerkin Neural Operator (GNO)}
\label{sec:gno}
To complement spectral and latent neural operators, we further consider a Galerkin Neural Operator (GNO)~\cite{Cao2021}, implemented through a Galerkin-style transformer architecture, for learning the cross-physics mapping from the diffusion field to the wave field. Unlike Fourier- or wavelet-based operators, which parameterize the global interaction through predefined basis functions, GNO represents the input field as a set of spatial--temporal tokens and learns global interactions through a linear Galerkin attention mechanism.
Given a normalized diffusion field
$
\mathbf{u}_{\mathrm{D}}
\in
\mathbb{R}^{1\times N_t\times N_x},
$
the GNO approximates the mapping
$
\mathcal{G}_{\theta}^{\mathrm{GNO}}:
\mathbf{u}_{\mathrm{D}}
\mapsto
\widehat{\mathbf{u}}_{\mathrm{W}},
\widehat{\mathbf{u}}_{\mathrm{W}}
=
\mathcal{G}_{\theta}^{\mathrm{GNO}}
\left(
\mathbf{u}_{\mathrm{D}}
\right),
$
where $N_t=501$ and $N_x=200$. The essential distinction is that the operator acts on a tokenized representation of the physical field, allowing global temporal--spatial interactions to be modeled without explicitly constructing dense pairwise attention matrices.

\paragraph{Patch embedding and tokenization.}
The input field is first partitioned into non-overlapping temporal--spatial patches. Because the temporal dimension $N_t=501$ is not divisible by the patch size, replication padding is first applied,
$
501\times200
\rightarrow
504\times200.
$
We use a patch size of
$
P_t\times P_x
=
4\times4.
$
A strided convolution with kernel size and stride equal to the patch size maps the padded diffusion field into a latent patch representation,
$
\mathbf{Z}_0
=
\mathcal{P}
\left(
\mathbf{u}_{\mathrm{D}}
\right)
\in
\mathbb{R}^{d\times H_p\times W_p},
$
where $d=96$ is the embedding dimension and
$
H_p
=
504/4
=
126,
W_p
=
200/4
=
50.
$
The resulting patch field therefore contains
$
N
=
H_pW_p
=
6300
$
tokens. After flattening the patch grid, the representation becomes
$
\mathbf{X}_0
\in
\mathbb{R}^{N\times d}.
$

\paragraph{Coordinate embedding.}
Since patch tokenization alone does not explicitly encode the physical location of each token, normalized temporal and spatial coordinates are added to the token representation. For each token location, we define
$
\boldsymbol{\xi}_{i}
=
\left[
\widetilde{t}_{i},
\widetilde{x}_{i}
\right],
\widetilde{t}_{i},
\widetilde{x}_{i}
\in[-1,1].
$
The coordinate pair is projected through a two-layer multilayer perceptron,
$
\mathbf{e}_{i}
=
\mathcal{E}_{\mathrm{coord}}
\left(
\boldsymbol{\xi}_{i}
\right)
\in
\mathbb{R}^{d},
$
and added to the corresponding patch token,
$
\widetilde{\mathbf{x}}_{i}
=
\mathbf{x}_{i}
+
\mathbf{e}_{i}.
$
This positional embedding allows the learned operator to retain the physical ordering of temporal--spatial patches.

\paragraph{Galerkin attention.}
The core of GNO is a Galerkin-style linear attention operator. Given a token matrix
$
\mathbf{X}
\in
\mathbb{R}^{N\times d},
$
query, key, and value features are constructed through learned linear projections,
$
\mathbf{Q}
=
\mathbf{X}\mathbf{W}_{Q},
\mathbf{K}
=
\mathbf{X}\mathbf{W}_{K},
\mathbf{V}
=
\mathbf{X}\mathbf{W}_{V}.
$
In contrast to standard self-attention,
$
\operatorname{Attention}
(\mathbf{Q},\mathbf{K},\mathbf{V})
=
\operatorname{softmax}
\left(
\mathbf{Q}\mathbf{K}^{\top}
/
\sqrt{d}
\right)
\mathbf{V},
$
the Galerkin operator does not form the $N\times N$ attention matrix. Instead, it evaluates
$
\mathcal{A}_{\mathrm{G}}
(\mathbf{Q},\mathbf{K},\mathbf{V})
=
\mathbf{Q}
\left(
\mathbf{K}^{\top}\mathbf{V}
/
N
\right).
$
The contraction
$
\mathbf{K}^{\top}\mathbf{V}
/
N
$
can be interpreted as a learned finite-dimensional projection or quadrature-like aggregation of the field features. The query features subsequently evaluate this global representation at each token position.
This formulation avoids explicitly constructing pairwise interactions between all token pairs. For token dimension $N$ and channel dimension $d$, standard attention requires an intermediate $N\times N$ matrix, whereas the Galerkin formulation contracts over the token dimension first and forms a $d\times d$ operator. This is particularly advantageous in the present setting, where the $4\times4$ patching still produces $N=6300$ tokens.

\paragraph{Multi-head Galerkin operator.}
We employ a multi-head formulation with $H=4$ heads. The embedding dimension is divided as
$
d_h
=
d/H
=
24.
$
For each head $h$, the Galerkin operator is evaluated independently,
$
\mathbf{Y}^{(h)}
=
\mathbf{Q}^{(h)}
\left[
\left(
\mathbf{K}^{(h)}
\right)^{\top}
\mathbf{V}^{(h)}
/
N
\right].
$
The head outputs are concatenated and projected back to the $d$-dimensional token space,
$
\mathbf{Y}
=
\mathcal{W}_{O}
\left(
\operatorname{Concat}
[
\mathbf{Y}^{(1)},
\ldots,
\mathbf{Y}^{(H)}
]
\right).
$
Layer normalization is applied independently to the query, key, and value head features before the Galerkin contraction. This feature normalization improves the conditioning of the global operator without introducing a softmax normalization over token pairs.

\paragraph{Galerkin operator block.}
Each operator block follows a pre-normalized residual architecture. Given token features $\mathbf{X}_{\ell}$, the Galerkin update is
$
\mathbf{X}_{\ell}^{\prime}
=
\mathbf{X}_{\ell}
+
\mathcal{A}_{\ell}^{\mathrm{G}}
\left(
\mathcal{N}_{1}
\left(
\mathbf{X}_{\ell}
\right)
\right),
$
where $\mathcal{N}_{1}$ denotes layer normalization. A feed-forward network is then applied through a second residual path,
$
\mathbf{X}_{\ell+1}
=
\mathbf{X}_{\ell}^{\prime}
+
\mathcal{F}_{\ell}
\left(
\mathcal{N}_{2}
\left(
\mathbf{X}_{\ell}^{\prime}
\right)
\right).
$
The feed-forward module consists of two linear layers with a GELU activation,
$
d
\rightarrow
2d
\rightarrow
d,
$
corresponding to an MLP expansion ratio of $2$.
We stack four Galerkin operator blocks,
$
\mathbf{X}_{\ell+1}
=
\mathcal{B}_{\ell}^{\mathrm{G}}
\left(
\mathbf{X}_{\ell}
\right),
\ell=0,\ldots,3.
$
The resulting token sequence therefore undergoes repeated global operator updates while preserving a fixed embedding width of $96$.

\paragraph{Latent field refinement.}
After the final Galerkin block, layer normalization is applied and the token sequence is reshaped back into a two-dimensional latent patch field,
$
\mathbf{Z}_{G}
\in
\mathbb{R}^{96\times126\times50}.
$
Although Galerkin attention captures global token interactions, patch tokenization and linear attention alone may not fully preserve local field continuity. We therefore introduce a residual convolutional refinement module,
$
\widetilde{\mathbf{Z}}_{G}
=
\phi
\left[
\mathbf{Z}_{G}
+
\mathcal{R}_{\mathrm{local}}
\left(
\mathbf{Z}_{G}
\right)
\right],
$
where $\mathcal{R}_{\mathrm{local}}$ contains two $3\times3$ convolutions with batch normalization, and $\phi(\cdot)$ is the GELU activation. This stage complements the global Galerkin interaction by restoring local temporal--spatial consistency on the latent patch grid.

\paragraph{Field reconstruction.}
The refined latent field is decoded to the full-resolution wave field. A transposed convolution with kernel size and stride equal to the patch size first performs patch-wise upsampling,
$
96\times126\times50
\rightarrow
64\times504\times200.
$
This is followed by two local convolutions,
$
64
\rightarrow
32
\rightarrow
1,
$
yielding
$
\widetilde{\mathbf{u}}_{\mathrm{W}}
\in
\mathbb{R}^{1\times504\times200}.
$
The padded temporal samples are finally removed,
$
\widehat{\mathbf{u}}_{\mathrm{W}}
=
\operatorname{Crop}
\left(
\widetilde{\mathbf{u}}_{\mathrm{W}}
\right)
\in
\mathbb{R}^{1\times501\times200}.
$
The complete architecture can therefore be summarized as
\begin{equation}
	\mathbf{u}_{\mathrm{D}}
	\xrightarrow{\mathrm{Patch}}
	\mathbf{X}_{\mathrm{D}}
	\xrightarrow{\mathrm{Galerkin\ operator}}
	\mathbf{X}_{\mathrm{W}}
	\xrightarrow{\mathrm{Local\ refinement}}
	\mathbf{Z}_{\mathrm{W}}
	\xrightarrow{\mathrm{Decoder}}
	\widehat{\mathbf{u}}_{\mathrm{W}}.
\end{equation}

\paragraph{Training objective.}
For consistency across all compared models, diffusion and wave fields are independently standardized using statistics computed exclusively from the training set. GNO is optimized using the same composite objective,
\begin{equation}
	\begin{split}
	\mathcal{L}_{\mathrm{GNO}}
	&=
	\mathcal{L}_{\mathrm{MSE}}
	+
	\lambda_{\mathrm{rel}}
	\mathcal{L}_{\mathrm{rel}},
	\qquad
	\lambda_{\mathrm{rel}}=0.1,
	\quad
	\\
	\mathcal{L}_{\mathrm{rel}}
	&=
	\frac{
		\left\|
		\widehat{\mathbf{u}}_{\mathrm{W}}
		-
		\mathbf{u}_{\mathrm{W}}
		\right\|_2
	}{
		\left\|
		\mathbf{u}_{\mathrm{W}}
		\right\|_2+\epsilon
	},
	\qquad
	\epsilon=10^{-8}.
	\end{split}
	\label{eq:gno_loss}
\end{equation}

Optimization is performed with AdamW using an initial learning rate of $10^{-3}$ and a weight decay of $10^{-5}$. A cosine learning-rate schedule is applied during training, the gradient norm is clipped at $1.0$, and the checkpoint with the minimum validation relative $\ell_2$ error is retained for final evaluation.
The key distinction between GNO and the preceding operator architectures lies in the form of the global interaction. U-NO and LNO parameterize nonlocal mappings through truncated Fourier modes, while WNO performs localized multiscale mixing in a wavelet basis. GNO instead constructs a learned global projection directly in feature space through the Galerkin contraction in Eq.~\eqref{eq:gno_loss}. It therefore provides a basis-independent token-space operator for testing whether cross-physics diffusion-to-wave transformations are better captured through learned projection interactions rather than predefined spectral representations.

\subsection{Cross-Physics Mapping Based on DeepONet}
\label{sec:deeponet}
To investigate cross-physics learning from the perspective of explicit operator decomposition, we further employ DeepONet~\cite{lu2019deeponet} to approximate the mapping from the diffusion field to the wave field. In contrast to convolutional and spectral neural operators, DeepONet represents the target operator through two complementary subnetworks: a branch network that encodes the input function and a trunk network that parameterizes the dependence of the output on the query coordinates. This decomposition provides a direct function-to-function representation of the diffusion-to-wave transformation.
Let
$
\mathbf{u}_{\mathrm{D}}
\in
\mathbb{R}^{1\times N_t\times N_x}
$
denote the normalized diffusion field, with $N_t=501$ and $N_x=200$. DeepONet approximates the cross-physics operator
$
\mathcal{G}_{\theta}^{\mathrm{DeepONet}}:
\mathbf{u}_{\mathrm{D}}
\mapsto
\widehat{\mathbf{u}}_{\mathrm{W}},
$
by decomposing the predicted wave field into input-dependent coefficients and coordinate-dependent basis functions.

\paragraph{Operator decomposition.}
For a query coordinate
$
\boldsymbol{\xi}
=
(t,x),
$
the DeepONet prediction is written as
$
\widehat{u}_{\mathrm{W}}
(\boldsymbol{\xi})
=
\frac{1}{\sqrt{p}}
\sum_{k=1}^{p}
b_k
\left(
\mathbf{u}_{\mathrm{D}}
\right)
\phi_k
\left(
\boldsymbol{\xi}
\right)
+
b_0,
$
where $p$ is the latent dimension,
$b_k(\mathbf{u}_{\mathrm{D}})$ are the coefficients produced by the branch network,
$\phi_k(\boldsymbol{\xi})$ are the coordinate-dependent basis functions generated by the trunk network, and $b_0$ is a learned scalar bias. We set
$
p=128.
$
The above equation makes the operator structure explicit: the branch network determines how the input diffusion field activates a set of latent modes, while the trunk network determines how these modes are evaluated over the temporal--spatial domain. The predicted wave field is obtained by combining the two representations through a latent inner product.

\paragraph{Branch network.}
The branch network receives the complete diffusion field,
$
\mathbf{u}_{\mathrm{D}}
\in
\mathbb{R}^{1\times501\times200},
$
and maps it to a global latent coefficient vector
$
\mathbf{b}
=
\mathcal{B}_{\theta_b}
\left(
\mathbf{u}_{\mathrm{D}}
\right)
\in
\mathbb{R}^{p}.
$
In our implementation, the branch encoder consists of five convolutional stages with channel dimensions
$
1
\rightarrow
32
\rightarrow
64
\rightarrow
128
\rightarrow
256
\rightarrow
256.
$
The first convolution uses a $5\times5$ kernel with stride $2$, while the subsequent layers use $3\times3$ kernels with stride $2$. Each convolution is followed by batch normalization and GELU activation. The corresponding temporal--spatial resolution is progressively reduced approximately as
$
501\times200
\rightarrow
251\times100
\rightarrow
126\times50
\rightarrow
63\times25
\rightarrow
32\times13
\rightarrow
16\times7.
$
The final convolutional representation is compressed using adaptive average pooling to a fixed $4\times4$ feature map,
$
\mathbf{Z}_{b}
\in
\mathbb{R}^{256\times4\times4}.
$
It is then flattened and projected through a two-layer multilayer perceptron,
$
256\times4\times4
\rightarrow
512
\rightarrow
128,
$
with GELU activation and a dropout probability of $0.1$. The resulting vector contains the global diffusion-dependent coefficients
$
\mathbf{b}
=
[
b_1,\ldots,b_p
].
$
The use of adaptive pooling decouples the final fully connected representation from the exact intermediate feature-map dimensions.

\paragraph{Trunk network.}
The trunk network represents the coordinate dependence of the output field. We construct a normalized temporal--spatial coordinate grid,
$
\boldsymbol{\xi}_{ij}
=
\left(
\widetilde{t}_i,
\widetilde{x}_j
\right),
\widetilde{t}_i,
\widetilde{x}_j
\in[-1,1],
$
for all
$
i=1,\ldots,N_t,
j=1,\ldots,N_x.
$
The grid contains
$
N_q
=
N_tN_x
=
501\times200
=
100200
$
query locations.
For each coordinate, the trunk network computes
$
\boldsymbol{\phi}
(\boldsymbol{\xi})
=
\mathcal{T}_{\theta_t}
(\boldsymbol{\xi})
\in
\mathbb{R}^{p},
$
where
$
\boldsymbol{\phi}
=
[
\phi_1,
\ldots,
\phi_p
].
$
The trunk is implemented as a multilayer perceptron with architecture
$
2
\rightarrow
128
\rightarrow
128
\rightarrow
128
\rightarrow
128
\rightarrow
128,
$
with GELU nonlinearities between successive layers. The final $128$-dimensional output acts as a learned basis representation associated with each temporal--spatial location.
Unlike the branch network, the trunk network depends only on the query coordinates and is therefore shared across all input diffusion fields. The branch and trunk networks play complementary roles: the former encodes the input-dependent global state of the diffusion field, whereas the latter defines how the learned latent representation is evaluated throughout the wave output domain.

\paragraph{Branch--trunk interaction.}
For a batch of diffusion fields, the branch network produces
$
\mathbf{B}
\in
\mathbb{R}^{B\times p},
$
while the trunk network produces
$
\mathbf{\Phi}
\in
\mathbb{R}^{N_q\times p}.
$
The complete output is obtained through the matrix contraction
$
\mathbf{Y}
=
\mathbf{B}
\mathbf{\Phi}^{\top}
/
\sqrt{p}
+
b_0,
$
where
$
\mathbf{Y}
\in
\mathbb{R}^{B\times N_q}.
$
The normalization factor $1/\sqrt{p}$ is introduced to control the initial numerical scale of the latent inner product. The resulting vector is then reshaped to the original field dimensions,
$
\widehat{\mathbf{u}}_{\mathrm{W}}
\in
\mathbb{R}^{B\times1\times501\times200}.
$
The complete DeepONet mapping can therefore be summarized as
\begin{equation}
	\mathbf{u}_{\mathrm{D}}
	\xrightarrow{\mathcal{B}_{\theta_b}}
	\mathbf{b}
	(t,x)
	\xrightarrow{\mathcal{T}_{\theta_t}}
	\boldsymbol{\phi}(t,x)
	\widehat{u}_{\mathrm{W}}(t,x)
	=
	\frac{
		\mathbf{b}^{\top}
		\boldsymbol{\phi}(t,x)
	}{
		\sqrt{p}
	}
	+b_0.
	\label{eq:deeponet_summary}
\end{equation}

\paragraph{Interpretation for cross-physics mapping.}
DeepONet imposes a fundamentally different inductive bias from the preceding architectures. ResUNet, U-NO, WNO, LNO, and GNO propagate or transform distributed feature fields, whereas DeepONet separates the mapping into an input-dependent coefficient representation and a coordinate-dependent basis representation. In this sense, the diffusion field is first compressed into a finite-dimensional set of coefficients,
$
\mathbf{u}_{\mathrm{D}}
\rightarrow
\mathbf{b},
$
and the corresponding wave response is reconstructed by evaluating these coefficients against learned basis functions over the output domain.
For the diffusion-to-wave problem, this decomposition tests whether the cross-physics operator admits an approximately low-rank representation of the form
$
\mathcal{G}
\left(
\mathbf{u}_{\mathrm{D}}
\right)
(t,x)
\approx
\sum_{k=1}^{p}
b_k
\left(
\mathbf{u}_{\mathrm{D}}
\right)
\phi_k(t,x).
$
Accordingly, the branch network learns how the full diffusion field determines the latent coefficients, while the trunk network learns a data-driven functional basis over the wave domain. This explicit separation between input-function encoding and output-coordinate evaluation provides a complementary operator-learning mechanism to the spectral, wavelet, latent, and Galerkin representations considered above.

\paragraph{Normalization and training objective.}
For consistency across all compared models, diffusion and wave fields are independently standardized using statistics computed exclusively from the training set,
$
\widetilde{\mathbf{u}}_{\mathrm{D}}
=
(\mathbf{u}_{\mathrm{D}}
-
\mu_{\mathrm{D}})
/
\sigma_{\mathrm{D}},
\widetilde{\mathbf{u}}_{\mathrm{W}}
=
(\mathbf{u}_{\mathrm{W}}
-
\mu_{\mathrm{W}})
/
\sigma_{\mathrm{W}}.
$
DeepONet is optimized using the same composite reconstruction objective,
\begin{equation}
	\begin{split}
	\mathcal{L}_{\mathrm{DeepONet}}
	&=
	\mathcal{L}_{\mathrm{MSE}}
	+
	\lambda_{\mathrm{rel}}
	\mathcal{L}_{\mathrm{rel}},
	\mathcal{L}_{\mathrm{MSE}}
	=
	\frac{1}{N}
	\left\|
	\widehat{\mathbf{u}}_{\mathrm{W}}
	-
	\mathbf{u}_{\mathrm{W}}
	\right\|_2^2,
	\\
	\mathcal{L}_{\mathrm{rel}}
	&=
	\frac{
		\left\|
		\widehat{\mathbf{u}}_{\mathrm{W}}
		-
		\mathbf{u}_{\mathrm{W}}
		\right\|_2
	}{
		\left\|
		\mathbf{u}_{\mathrm{W}}
		\right\|_2
		+
		\epsilon
	},
	\epsilon=10^{-8}.
	\end{split}
	\label{eq:deeponet_loss}
\end{equation}
where $\lambda_{\mathrm{rel}}=0.1$.
Optimization is performed using AdamW with an initial learning rate of $10^{-3}$ and weight decay of $10^{-5}$. A cosine learning-rate schedule is employed throughout training, and the gradient norm is clipped at $1.0$. The checkpoint yielding the minimum validation relative $\ell_2$ error is retained for final testing.
The principal distinction of DeepONet is therefore its explicit separation between ``what'' is encoded from the input field and ``where'' the output is evaluated. While Fourier- and wavelet-based neural operators construct transformations through prescribed spectral representations, and GNO learns global interactions directly between field tokens, DeepONet approximates the diffusion-to-wave operator through a learned separable expansion in which the branch network produces input-dependent coefficients and the trunk network produces coordinate-dependent basis functions. It thus provides an important complementary baseline for evaluating whether the cross-physics transformation admits a compact functional decomposition.

\subsection{Cross-Physics Mapping Based on Fourier Neural Operator (FNO)}
\label{sec:fno}
We finally consider the Fourier Neural Operator (FNO)~\cite{li2020fourier} as a canonical spectral operator baseline for learning the cross-physics transformation from the diffusion field to the wave field. In contrast to conventional convolutional networks, which propagate information through spatially local kernels, FNO parameterizes global interactions directly in the Fourier domain. This property makes it particularly suitable for cross-physics mappings in which the output at a given temporal--spatial location may depend on globally distributed structures in the input field.
Given a normalized diffusion field
$
\mathbf{u}_{\mathrm{D}}
\in
\mathbb{R}^{1\times N_t\times N_x},
$
with $N_t=501$ and $N_x=200$, FNO learns the operator
$
\mathcal{G}_{\theta}^{\mathrm{FNO}}:
\mathbf{u}_{\mathrm{D}}
\mapsto
\widehat{\mathbf{u}}_{\mathrm{W}},
\widehat{\mathbf{u}}_{\mathrm{W}}
=
\mathcal{G}_{\theta}^{\mathrm{FNO}}
\left(
\mathbf{u}_{\mathrm{D}}
\right).
$
The operator is constructed by alternating global Fourier-domain mixing and local pointwise transformations in a lifted latent feature space.

\paragraph{Coordinate lifting.}
To retain explicit information about the temporal--spatial location of each field value, the diffusion input is augmented with normalized coordinates,
$
\mathbf{g}(t,x)
=
\left[
\widetilde{t},
\widetilde{x}
\right],
\widetilde{t},
\widetilde{x}
\in[0,1].
$
The input representation is therefore
$
\mathbf{v}_0
=
\operatorname{Concat}
\left[
\mathbf{u}_{\mathrm{D}},
\widetilde{t},
\widetilde{x}
\right],
$
which contains three channels. A pointwise lifting operator maps this representation into a latent field,
$
\mathbf{z}_0
=
\mathcal{P}
\left(
\mathbf{v}_0
\right),
\mathbf{z}_0
\in
\mathbb{R}^{C\times N_t\times N_x},
$
where $\mathcal{P}$ is implemented as a $1\times1$ convolution and the latent width is set to
$
C=64.
$

\paragraph{Fourier integral operator.}
The central component of FNO is a spectral convolution that parameterizes a global integral operator in Fourier space. For a latent feature field $\mathbf{z}_{\ell}$, the two-dimensional Fourier transform over the temporal and spatial dimensions is
$
\widehat{\mathbf{z}}_{\ell}
=
\mathcal{F}
\left(
\mathbf{z}_{\ell}
\right).
$
Instead of learning a dense kernel directly in the physical domain, FNO learns complex-valued transformations on a truncated set of Fourier coefficients,
$
\widehat{\mathbf{v}}_{\ell}
(k_t,k_x)
=
\mathbf{R}_{\ell}
(k_t,k_x)
\widehat{\mathbf{z}}_{\ell}
(k_t,k_x),
(k_t,k_x)\in\mathcal{M},
$
where
$\mathbf{R}_{\ell}(k_t,k_x)$
is a trainable complex-valued tensor and
$\mathcal{M}$ denotes the retained Fourier modes.
The output of the spectral operator is reconstructed through the inverse Fourier transform,
$
\mathcal{K}_{\ell}
\left(
\mathbf{z}_{\ell}
\right)
=
\mathcal{F}^{-1}
\left[
\widehat{\mathbf{v}}_{\ell}
\right].
$
Because Fourier basis functions have global support, each spectral layer can propagate information across the entire temporal--spatial field within a single operator update.
In our implementation, the retained spectral bandwidth is
$
(M_t,M_x)
=
(96,48),
$
and the actual number of modes is clipped automatically when it exceeds the available spectral resolution. The implementation uses the real-valued multidimensional FFT, such that only the nonredundant spatial-frequency coefficients are explicitly represented along the final transformed dimension.

\paragraph{Fourier operator block.}
Each FNO block combines the global spectral operator with a local pointwise transformation. Given $\mathbf{z}_{\ell}$, the update is
$
\mathbf{z}_{\ell+1}
=
\phi
\left[
\mathcal{K}_{\ell}
\left(
\mathbf{z}_{\ell}
\right)
+
\mathcal{W}_{\ell}
\left(
\mathbf{z}_{\ell}
\right)
\right],
\label{eq:fno_block}
$
where
$\mathcal{W}_{\ell}$
is a learned $1\times1$ convolution and
$\phi(\cdot)$
denotes the GELU activation.
The two branches have complementary roles. The spectral branch models nonlocal dependencies through Fourier-space multiplication, whereas the pointwise branch directly transforms local channel features. Their combination enables the network to represent both long-range operator interactions and local corrections.
We use four consecutive Fourier operator blocks,
$
\mathbf{z}_{\ell+1}
=
\mathcal{B}_{\ell}^{\mathrm{FNO}}
\left(
\mathbf{z}_{\ell}
\right),
\ell=0,\ldots,3,
$
while keeping the latent width fixed at $64$ throughout the operator layers.

\paragraph{Output projection.}
After the final spectral block, the latent field is mapped back to the wave domain through two pointwise projection layers,
$
\widehat{\mathbf{u}}_{\mathrm{W}}
=
\mathcal{Q}_2
\left[
\phi
\left(
\mathcal{Q}_1
\left(
\mathbf{z}_4
\right)
\right)
\right],
$
where
$
64
\rightarrow
128
\rightarrow
1.
$
Since the spectral blocks preserve the temporal--spatial resolution, no encoder--decoder reconstruction or spatial interpolation is required, and the final prediction directly retains the original resolution,
$
\widehat{\mathbf{u}}_{\mathrm{W}}
\in
\mathbb{R}^{1\times501\times200}.
$
The complete FNO architecture can therefore be summarized as
\begin{equation}
	\mathbf{u}_{\mathrm{D}}
	\xrightarrow{\mathrm{Coordinate\ lifting}}
	\mathbf{z}_0
	\xrightarrow{
		4\times\mathrm{Fourier\ operator}
	}
	\mathbf{z}_4
	\xrightarrow{\mathrm{Projection}}
	\widehat{\mathbf{u}}_{\mathrm{W}}.
	\label{eq:fno_summary}
\end{equation}

\paragraph{Wave-aware training objective.}
While the preceding models are primarily trained using field reconstruction errors, the oscillatory nature of the wave target motivates an additional constraint on the local wavefront structure. We therefore augment the reconstruction objective with temporal and spatial gradient discrepancies.
For a discrete wave field $\mathbf{u}$, the first-order temporal and spatial differences are defined as
$
\nabla_t
\mathbf{u}_{i,j}
=
\mathbf{u}_{i+1,j}
-
\mathbf{u}_{i,j},
$
and
$
\nabla_x
\mathbf{u}_{i,j}
=
\mathbf{u}_{i,j+1}
-
\mathbf{u}_{i,j}.
$
The corresponding gradient losses are
\begin{equation}
	\mathcal{L}_{t}
	=
	\left\|
	\nabla_t
	\widehat{\mathbf{u}}_{\mathrm{W}}
	-
	\nabla_t
	\mathbf{u}_{\mathrm{W}}
	\right\|_1,
	\quad
	\mathcal{L}_{x}
	=
	\left\|
	\nabla_x
	\widehat{\mathbf{u}}_{\mathrm{W}}
	-
	\nabla_x
	\mathbf{u}_{\mathrm{W}}
	\right\|_1.
\end{equation}

These terms encourage the predicted field to reproduce temporal transitions and spatial wavefront structures that may be insufficiently constrained by pointwise reconstruction alone.
The total training objective is
\begin{equation}
	\mathcal{L}_{\mathrm{FNO}}
	=
	\mathcal{L}_{\mathrm{MSE}}
	+
	\lambda_{\mathrm{rel}}
	\mathcal{L}_{\mathrm{rel}}
	+
	\lambda_t
	\mathcal{L}_{t}
	+
	\lambda_x
	\mathcal{L}_{x},
	\mathcal{L}_{\mathrm{rel}}
	=
	\frac{
		\left\|
		\widehat{\mathbf{u}}_{\mathrm{W}}
		-
		\mathbf{u}_{\mathrm{W}}
		\right\|_2
	}{
		\left\|
		\mathbf{u}_{\mathrm{W}}
		\right\|_2
		+
		\epsilon
	}.
	\label{eq:fno_loss}
\end{equation}

We use
$
\lambda_{\mathrm{rel}}
=
0.1,
\lambda_t
=
\lambda_x
=
0.02,
\epsilon
=
10^{-8}.
$
We additionally implemented a Fourier-domain discrepancy,
\begin{equation}
	\mathcal{L}_{\mathrm{FFT}}
	=
	\frac{1}{N}
	\left\|
	\mathcal{F}
	\left(
	\widehat{\mathbf{u}}_{\mathrm{W}}
	\right)
	-
	\mathcal{F}
	\left(
	\mathbf{u}_{\mathrm{W}}
	\right)
	\right\|_1,
	\label{eq:fno_fft_loss}
\end{equation}
which allows spectral agreement to be explicitly incorporated into the objective. In the experiments reported here, however, its weight is set to
$
\lambda_{\mathrm{FFT}}=0,
$
such that the final optimized objective is given by Eq.~\eqref{eq:fno_loss}. This choice isolates the effect of the temporal--spatial gradient regularization from an additional spectral-domain supervision term.

\paragraph{Normalization and optimization.}
The diffusion and wave fields are independently standardized using statistics computed exclusively from the training set,
$
\widetilde{\mathbf{u}}_{\mathrm{D}}
=
(\mathbf{u}_{\mathrm{D}}
-
\mu_{\mathrm{D}})
/
\sigma_{\mathrm{D}},
\widetilde{\mathbf{u}}_{\mathrm{W}}
=
(\mathbf{u}_{\mathrm{W}}
-
\mu_{\mathrm{W}})
/
\sigma_{\mathrm{W}}.
$
Optimization is performed using AdamW with an initial learning rate of $10^{-3}$ and weight decay of $10^{-5}$. A cosine learning-rate schedule is applied during training, and the gradient norm is clipped at $1.0$. The checkpoint with the minimum validation relative $\ell_2$ error is retained for final evaluation.
From the perspective of cross-physics operator learning, FNO provides the most direct spectral representation among the architectures considered here. ResUNet constructs global context indirectly through stacked local convolutions, U-NO introduces Fourier operators within a multiresolution U-shaped hierarchy, LNO performs spectral mixing after dimensional compression, WNO replaces global Fourier modes with localized wavelet coefficients, GNO learns global projection interactions in token space, and DeepONet factorizes the operator into input-dependent coefficients and coordinate-dependent basis functions. FNO instead retains the full temporal--spatial field resolution and repeatedly applies global Fourier-domain transformations. It therefore serves as a canonical reference for assessing whether direct global spectral mixing is sufficient to learn the transformation from diffusive diffusion dynamics to propagating wave dynamics.

\begin{figure*}[t]
	\centering
	\includegraphics[width=\textwidth]{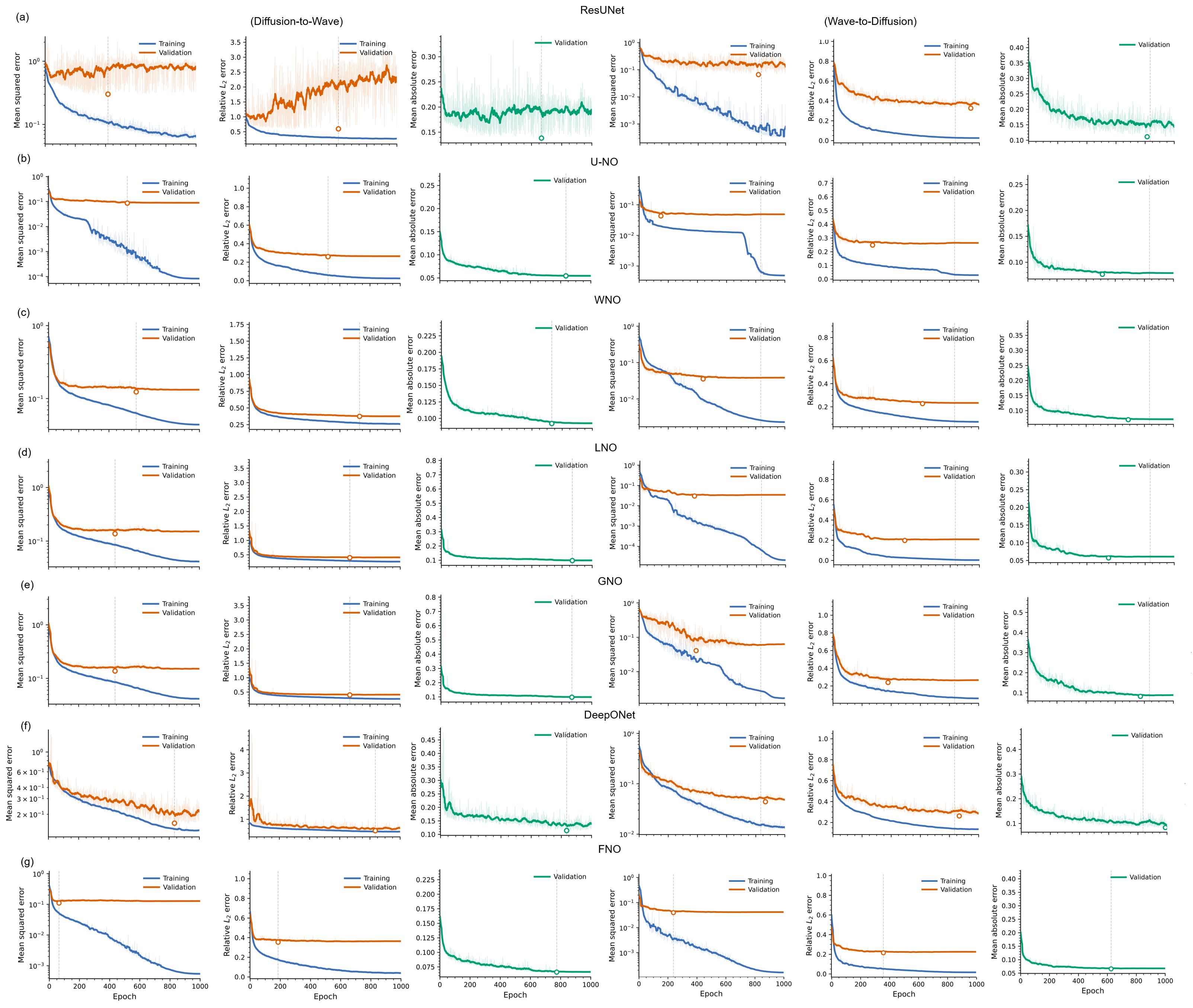}
	\caption{Loss curves of different neural networks.}
	\label{fig10}
\end{figure*}

\section{Results and Discussion}
\subsection{Diffusion-to-Wave}
Figure~\ref{fig10} compares the training dynamics of the seven considered architectures using the mean squared error (MSE), relative $L_2$ error, and mean absolute error (MAE). Clear differences are observed between the conventional convolutional baseline and the neural-operator-based models. In general, most neural operators exhibit a rapid error reduction during the initial stage of training, followed by a progressively slower convergence toward a stable validation plateau. This behavior indicates that the dominant components of the wave-to-diffusion mapping are learned relatively early, whereas subsequent optimization mainly refines higher-order or smaller-amplitude structures.
Among all architectures, U-NO shows the most favorable overall convergence and generalization behavior. Its training MSE decreases by several orders of magnitude, while the validation MSE rapidly approaches a stable level without noticeable late-stage degradation. A similar trend is observed for the relative $L_2$ error and MAE, for which the validation curves converge to approximately $0.27$ and $0.055$, respectively. Although a non-negligible gap remains between the training and validation errors, the validation curves are smooth and nearly monotonic, suggesting that the model captures a transferable operator rather than simply fitting individual training realizations.
FNO exhibits similarly strong performance, particularly in terms of absolute reconstruction accuracy. The training MSE continues to decrease throughout optimization and reaches the lowest range among the investigated models, while the validation MAE converges to approximately $0.065$. The validation relative $L_2$ error, however, saturates at a somewhat higher level than that of U-NO. The increasing separation between the training and validation curves at later epochs indicates mild overfitting, despite the overall stable validation performance. This result suggests that global Fourier-domain representations are highly effective at learning the dominant spatial-temporal correlations between the acoustic and thermal fields, but further reduction of the training residual does not necessarily translate into improved out-of-distribution accuracy.

The WNO, LNO, and GNO models display intermediate and relatively similar behavior. All three architectures undergo a sharp reduction in error during the first tens of epochs, after which the validation metrics gradually approach stable plateaus. WNO achieves slightly lower validation errors than LNO and GNO, with a final MAE of approximately $0.09$, whereas the latter two models converge to values close to $0.10$. Their relatively small temporal fluctuations and absence of pronounced validation-error growth indicate stable optimization. Nevertheless, their higher residual errors compared with U-NO and FNO imply that the corresponding wavelet, latent-space, and Galerkin representations provide a less efficient parameterization of the present cross-physics mapping under the same training conditions.
The behavior of DeepONet is qualitatively different. Although both training and validation errors decrease continuously, the convergence is slower and the curves exhibit substantially stronger fluctuations, particularly during the early and intermediate stages of training. The final MAE remains around $0.13\!-\!0.15$, higher than those obtained with the best neural operators. This may be associated with the branch-trunk decomposition of DeepONet, in which the input field and query coordinates are encoded separately. While this formulation is highly general for operator learning, it may be less efficient for the present problem, where the input and output are densely sampled spatiotemporal fields defined on the same structured domain.
In contrast, ResUNet shows pronounced overfitting and the weakest generalization among the tested architectures. Its training MSE decreases continuously, whereas the validation MSE remains large and highly fluctuating. More notably, the validation relative $L_2$ error increases during training even though the corresponding training error continues to decrease. This divergence demonstrates that improved fitting of the training samples does not lead to a better reconstruction of unseen fields. The result highlights an important distinction between conventional image-to-image regression and operator learning for the present cross-physics problem: local convolutional feature extraction alone is insufficient to robustly represent the global spatiotemporal transformation connecting propagative wave responses to diffusive heat fields.

Taken together, these results demonstrate that operator-learning architectures are substantially better suited to the wave-to-diffusion transformation than the conventional ResUNet baseline. In particular, U-NO provides the best balance between validation accuracy and training stability, while FNO achieves highly competitive performance with particularly low absolute reconstruction errors. The superior behavior of these architectures is consistent with the intrinsically nonlocal nature of the cross-physics mapping: the thermal response at a given space-time location is not determined solely by local acoustic features, but depends on information distributed over an extended spatiotemporal domain. Architectures capable of directly learning such global operators therefore provide a more appropriate inductive bias for translating wave-like measurements into diffusion-dominated fields.

Table~\ref{tab:model_comparison} quantitatively confirms the trends observed from the training and validation curves. Among all evaluated architectures, U-NO achieves the best performance across all five metrics, with an MSE of $1.02\times10^{10}$, an MAE of $2.18\times10^{4}$, an RMSE of $1.01\times10^{5}$, a relative $L_2$ error of 0.307, and an $R^2$ of 0.905. In particular, U-NO is the only model that reaches an $R^2$ above 0.9, indicating that it captures most of the variance in the target field while simultaneously maintaining the lowest absolute and relative reconstruction errors. Compared with FNO, which provides the second-best performance among the spectral/operator-learning models in several metrics, U-NO reduces the MSE by approximately $39\%$, the MAE by $33\%$, and the relative $L_2$ error by $21\%$. These results indicate that the multiscale U-shaped operator representation is particularly effective for the present wave-to-diffusion cross-physics transformation.
LNO provides the second-best overall performance, achieving an MSE of $1.44\times10^{10}$, a relative $L_2$ error of 0.365, and an $R^2$ of 0.866. FNO, WNO, and GNO form a second performance group, with relatively similar MSE values of $1.66\!-\!1.72\times10^{10}$, relative $L_2$ errors of approximately $0.39\!-\!0.40$, and $R^2$ values around $0.84$. The comparable performance of these three architectures suggests that Fourier-, wavelet-, and Galerkin-based representations are all capable of capturing the dominant nonlocal correlations underlying the cross-physics mapping, although their reconstruction accuracy remains below that of U-NO.

By contrast, DeepONet and ResUNet exhibit substantially larger reconstruction errors, with relative $L_2$ errors of 0.548 and 0.528 and $R^2$ values of 0.698 and 0.719, respectively. ResUNet therefore improves only marginally over DeepONet and remains clearly inferior to the neural-operator architectures. This comparison is particularly informative because it suggests that the performance gain is not simply associated with using a deep neural network, but rather with employing architectures that are explicitly designed to approximate mappings between function spaces. For the wave-to-diffusion problem, the output at a given space-time location depends on information distributed over an extended spatiotemporal domain; consequently, operator-learning models with global or multiscale receptive mechanisms are better suited to represent this intrinsically nonlocal transformation.
Taken together, Table~\ref{tab:model_comparison} establishes U-NO as the most accurate model for the present cross-physics mapping task, while the overall superiority of U-NO, LNO, FNO, WNO, and GNO over the conventional ResUNet baseline further supports the use of neural operators for learning the transformation between propagative acoustic fields and diffusion-dominated thermal fields.

\begin{table*}[t]
	\centering
	\caption{Performance comparison of different models for diffusion-to-wave mapping.}
	\label{tab:model_comparison}
	\renewcommand{\arraystretch}{1.15}
	\setlength{\tabcolsep}{6pt}
	\begin{tabular}{lccccc}
		\toprule
		\textbf{Model} &
		\textbf{MSE} &
		\textbf{MAE} &
		\textbf{RMSE} &
		\textbf{Rel. $L_2$} &
		\textbf{$R^2$} \\
		\midrule
		DeepONet
		& $3.25\times10^{10}$
		& $4.51\times10^{4}$
		& $1.80\times10^{5}$
		& 0.548
		& 0.698 \\
		
		FNO
		& $1.66\times10^{10}$
		& $3.23\times10^{4}$
		& $1.29\times10^{5}$
		& 0.391
		& 0.846 \\
		
		ResUNet
		& $3.03\times10^{10}$
		& $4.43\times10^{4}$
		& $1.74\times10^{5}$
		& 0.528
		& 0.719 \\
		
		\textbf{U-NO}
		& $\mathbf{1.02\times10^{10}}$
		& $\mathbf{2.18\times10^{4}}$
		& $\mathbf{1.01\times10^{5}}$
		& \textbf{0.307}
		& \textbf{0.905} \\
		
		WNO
		& $1.72\times10^{10}$
		& $3.82\times10^{4}$
		& $1.31\times10^{5}$
		& 0.398
		& 0.841 \\
		
		GNO
		& $1.71\times10^{10}$
		& $3.57\times10^{4}$
		& $1.31\times10^{5}$
		& 0.397
		& 0.841 \\
		
		LNO
		& $1.44\times10^{10}$
		& $2.69\times10^{4}$
		& $1.20\times10^{5}$
		& 0.365
		& 0.866 \\
		\bottomrule
	\end{tabular}
\end{table*}

\begin{figure*}[t]
	\centering
	\includegraphics[width=\textwidth]{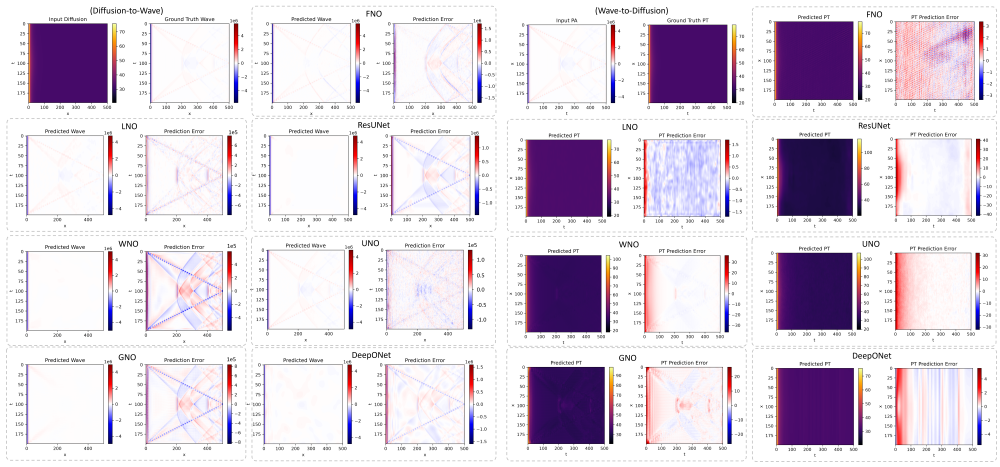}
	\caption{Qualitative comparison of diffusion-to-wave (left) and wave-to-diffusion (right) cross-physics reconstruction for a plate containing a square-hole defect under a $0.01\,\mathrm{s}$ short-pulse excitation.
		The diffusion/wave input, ground-truth wave/diffusion field, and predictions obtained using FNO, LNO, ResUNet, WNO, U-NO, GNO, and DeepONet are shown together with their corresponding prediction errors.
	}
	\label{fig11}
\end{figure*}

Figure~\ref{fig11} provides a representative qualitative comparison for a plate containing a square-hole defect subjected to a $0.01\,\mathrm{s}$ short-pulse excitation. This example is particularly informative because the input diffusion field is strongly smoothed and contains little explicit information resembling wave propagation, whereas the corresponding target field exhibits distinct diagonal wavefronts, reflections, and defect-induced scattering features. The reconstruction therefore requires the network to infer a highly nonlocal spatiotemporal transformation rather than perform a simple pointwise or local image-to-image conversion.
Clear differences are observed among the investigated architectures. U-NO provides the closest visual agreement with the ground-truth wave field, reproducing both the principal propagating branches and the secondary structures generated by interaction with the square-hole defect. Its error map is comparatively weak and largely unstructured, indicating that the remaining discrepancy is distributed over small-amplitude components rather than being concentrated along the dominant wavefronts. This is consistent with the quantitative results, in which U-NO achieves the lowest reconstruction errors and the highest $R^2$.
LNO also reconstructs the main spatiotemporal topology of the wave field with good fidelity. The principal incident and scattered branches are preserved, although its residual map still contains coherent errors along some wavefront trajectories, particularly around the region where the propagating field interacts with the defect. This suggests that the latent representation captures most of the global dynamics but does not completely resolve the finer-amplitude and phase-dependent components of the wave response.
FNO successfully recovers the dominant wave-propagation pattern, including the principal diagonal branches and a substantial portion of the defect-associated structure. However, its residual is noticeably more structured than those of U-NO and LNO. The remaining error follows several propagating branches, indicating small discrepancies in wavefront amplitude, phase, or arrival-time reconstruction. This observation is consistent with the fact that a small spatial or temporal shift of a sharp wavefront can produce a relatively large pointwise residual even when the overall field topology is correctly reproduced.
The limitations of the other architectures are more apparent in this example. ResUNet reconstructs only a weak fraction of the target response, while its error map largely retains the characteristic wavefront geometry of the ground truth. This indicates that the network fails to generate substantial portions of the propagating field rather than merely introducing small local inaccuracies. A similar behavior is observed for DeepONet, whose prediction is dominated by low-amplitude and spatially smooth components; consequently, the residual contains pronounced incident, reflected, and scattered wavefronts. These results suggest that the corresponding representations are less effective at recovering the strongly nonlocal and highly structured dynamics required by the diffusion-to-wave transformation.
WNO and GNO show intermediate behavior. Both models reproduce part of the large-scale wave response, but coherent residuals remain concentrated along the main wavefronts and around the defect-interaction region. The persistence of these structured residuals indicates that the global propagation geometry is only partially recovered and that the reconstruction of finer scattering features remains challenging.
More importantly, this example illustrates the fundamental difficulty of the cross-physics mapping. The diffusion field has undergone strong spatial and temporal smoothing, whereas the target wave field contains sharp propagation fronts and rich time-of-flight information. Therefore, successful reconstruction requires the learned operator to recover organized wave-like structures from a representation in which such features are not directly visible. The superior performance of U-NO indicates that its multiscale operator architecture provides an effective inductive bias for simultaneously capturing long-range propagation and localized defect interactions. The visual comparison thus supports the quantitative benchmark and further demonstrates that the advantage of neural operators lies not only in reducing scalar reconstruction errors, but also in preserving the physically meaningful spatiotemporal structure of the reconstructed field.

\begin{figure*}[t]
	\centering
	\includegraphics[width=\textwidth]{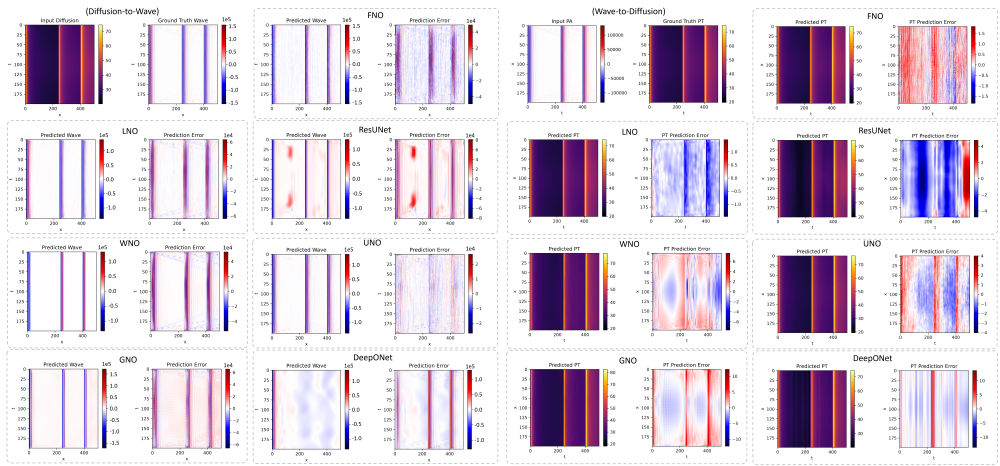}
	\caption{Qualitative comparison of diffusion-to-wave (left) and wave-to-diffusion (right) reconstruction for a defect-free plate under chirp excitation.
	The diffusion/wave input, ground-truth wave/diffusion field, and predictions obtained using FNO, LNO, ResUNet, WNO, U-NO, GNO, and DeepONet are shown together with the corresponding prediction errors.
	}
	\label{fig12}
\end{figure*}

Figure~\ref{fig12} further evaluates the models using a defect-free plate subjected to chirp excitation. In contrast to the short-pulse defective case, this configuration removes defect-induced scattering and therefore provides a cleaner assessment of whether the learned operators can reproduce the intrinsic cross-physics correspondence generated by a broadband excitation. The ground-truth wave field is dominated by several coherent high-amplitude structures that persist over the observation window, together with weaker distributed components. Successful reconstruction consequently requires accurate preservation of their spatial locations, amplitudes, signs, and temporal coherence.
Among the investigated architectures, U-NO again provides the closest agreement with the ground truth. The dominant wave-field components are reconstructed at the correct spatial positions and with amplitudes closely matching the reference solution. More importantly, the corresponding prediction error is relatively weak and lacks pronounced coherent structures. Since systematic errors in wave-field reconstruction generally appear as spatially or temporally organized residuals, the largely unstructured error produced by U-NO indicates that both the dominant field components and their temporal evolution are accurately represented.
FNO also reproduces the main features of the chirp-driven response with high fidelity. The principal high-amplitude components are correctly localized and remain coherent over time. Its error map, however, exhibits stronger vertically organized residuals around the dominant field structures, suggesting small but systematic discrepancies in their amplitude and/or phase. These errors are considerably smaller than the magnitude of the reconstructed field, but their structured nature indicates that some components of the broadband response are not completely resolved.
LNO shows a similar overall reconstruction capability, correctly recovering the dominant spatial structure of the wave field. Nevertheless, coherent residuals remain concentrated around the major response components, with additional weak errors distributed throughout the spatiotemporal domain. This suggests that the latent representation retains the principal cross-physics information but introduces small reconstruction biases when recovering the detailed broadband response.
WNO and GNO also capture the overall spatial organization of the target field, but their error maps contain more pronounced coherent patterns. In particular, the residuals are strongly concentrated along the high-amplitude field components, indicating that these models identify the correct locations of the principal response but reproduce their amplitudes and fine temporal characteristics less accurately. Thus, their errors arise predominantly from imperfect reconstruction of existing physical structures rather than from a complete failure to identify the global response.
The limitations of ResUNet and DeepONet are considerably more apparent. Although ResUNet identifies some of the dominant spatial components, it additionally generates broad, low-frequency structures that are absent from the ground truth. These artifacts lead to large and spatially correlated residuals, indicating that conventional convolutional feature extraction does not adequately preserve the global relationship between the diffusion and wave fields under broadband excitation. DeepONet exhibits an even stronger smoothing effect: apart from the strongest response components, much of the weaker wave-field structure is suppressed and replaced by slowly varying patterns. Consequently, its error map retains a substantial fraction of the ground-truth response.
Importantly, the defect-free chirp case demonstrates that the superior performance of the operator-learning models is not dependent on the presence of defect-generated features. Even when the response is governed primarily by the excitation and propagation characteristics of the pristine medium, U-NO, FNO, and LNO maintain a clear advantage. This indicates that these architectures learn the underlying field-to-field transformation itself rather than merely recognizing characteristic defect signatures.

\begin{figure*}[t]
	\centering
	\includegraphics[width=\textwidth]{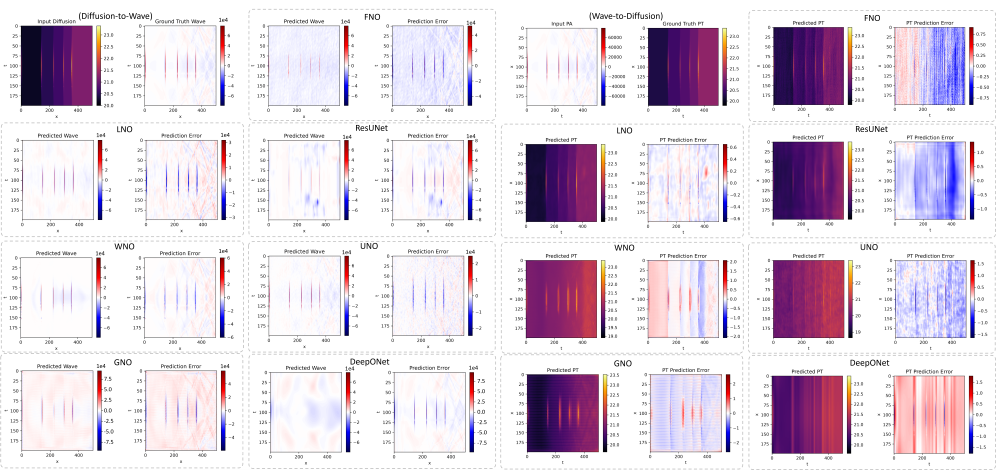}
	\caption{Qualitative comparison of diffusion-to-wave (left) and wave-to-diffusion (right) reconstruction for the case of a centered Gaussian-chirp spatially nonuniform excitation.
	The diffusion/wave input, ground-truth wave/diffusion field, and predictions obtained using FNO, LNO, ResUNet, WNO, U-NO, GNO, and DeepONet are shown together with their corresponding prediction errors.
	}
	\label{fig13}
\end{figure*}

Figure~\ref{fig13} presents a representative qualitative comparison for the case of a centered Gaussian-chirp spatially nonuniform excitation. In contrast to the previous uniform-excitation examples, this configuration introduces an additional level of difficulty because the input diffusion field is no longer spatially balanced. Instead, the excitation produces a nonuniform energy distribution, which leads in the target wave field to several localized high-amplitude response peaks together with weaker distributed background components. Successful reconstruction therefore requires the model to recover not only the positions of the dominant wave features, but also their relative amplitudes and the weaker surrounding field structure under a nonuniform excitation condition.
Among all investigated models, U-NO again provides the best qualitative agreement with the ground-truth wave field. It accurately reconstructs the main localized wave responses and preserves their amplitudes and spatial positions with high fidelity. At the same time, the prediction error remains relatively weak and largely unstructured, indicating that the discrepancy is mainly confined to small residual fluctuations rather than systematic failure along the dominant wave components. This behavior is particularly important in the present case, because spatially nonuniform excitation can easily lead to biased amplitude reconstruction if the learned mapping does not properly account for global field interactions.
LNO also performs strongly in this example, reproducing the principal localized response peaks with good agreement to the ground truth. Its residual map, however, still exhibits coherent error patterns around the dominant peaks and in the background region, suggesting that while the main field structure is captured, some inaccuracies remain in the reconstruction of weaker components and fine amplitude variations. FNO shows comparable behavior, correctly identifying the major response locations and their general strengths, but with more visible structured residuals distributed across the spatiotemporal domain. These residuals indicate that the dominant response is recovered, although the detailed amplitude distribution and low-amplitude background structures are not reproduced as accurately as with U-NO.
WNO yields an intermediate reconstruction quality. It captures the principal wave-response peaks and part of the surrounding field, but introduces noticeable coherent residuals, especially in the vicinity of the dominant structures. This suggests that the model can recover the main response pattern but is less effective at representing the full multiscale structure induced by the nonuniform Gaussian-chirp excitation.
The remaining models show more pronounced limitations. ResUNet produces visible spurious low-frequency artifacts and smoothing effects, which distort both the dominant peaks and the weaker background field. DeepONet exhibits even stronger smoothing, suppressing much of the physically meaningful structure and replacing it with broad, slowly varying components. As a result, its error map retains a substantial portion of the true wave-field features. GNO also shows clear structured distortions, with nonphysical background patterns superimposed on the reconstructed field, indicating that it is less robust under this more heterogeneous excitation scenario.
This example further demonstrates that the relative ranking of the models remains consistent even when the excitation becomes spatially nonuniform. In particular, U-NO retains the strongest robustness across changing excitation conditions, while LNO and FNO remain competitive but less accurate in reconstructing fine-scale amplitude details. More broadly, the figure shows that the advantage of neural operators is not limited to recovering wavefront geometry alone; it also extends to preserving the correct amplitude organization of the wave field under nonuniform forcing. This is especially relevant for practical cross-physics mapping, where excitation conditions are often spatially heterogeneous and the ability to generalize across such variations is essential.

\begin{figure*}[t]
	\centering
	\includegraphics[width=\textwidth]{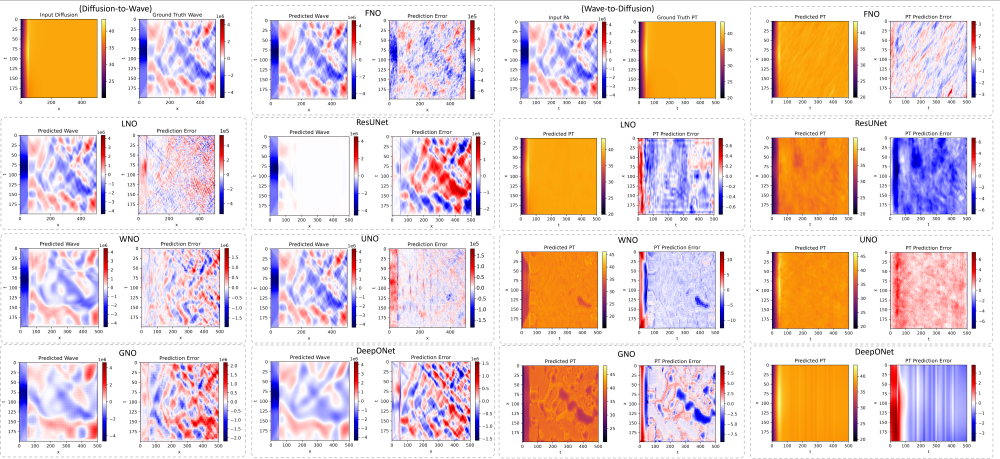}
	\caption{Qualitative comparison of diffusion-to-wave (left) and wave-to-diffusion (right) reconstruction for an asymmetric, spatially nonuniform localized Gaussian long-pulse excitation of $1\,\mathrm{s}$ duration.
	The diffusion/wave input, ground-truth wave/diffusion field, and predictions obtained using FNO, LNO, ResUNet, WNO, U-NO, GNO, and DeepONet are shown together with their corresponding prediction errors.
	}
	\label{fig14}
\end{figure*}

Figure~\ref{fig14} considers a more challenging excitation scenario consisting of an asymmetric, spatially nonuniform localized Gaussian pulse with a duration of $1\,\mathrm{s}$. Compared with the short-pulse and chirp cases discussed above, the longer excitation duration produces a substantially richer spatiotemporal response, characterized by broad coherent structures, spatially varying amplitudes, and complex superposition patterns across the observation domain. At the same time, the diffusion input remains comparatively smooth, making the recovery of the corresponding wave field a strongly nonlocal cross-physics reconstruction problem.
Among the investigated models, U-NO again achieves the closest qualitative agreement with the ground truth. It reproduces the dominant positive and negative field regions, their spatial organization, and a substantial fraction of the finer spatiotemporal textures visible in the reference wave field. Its prediction error is considerably weaker than those of the other models and is predominantly composed of small-amplitude, weakly organized fluctuations. The absence of large coherent structures in the residual indicates that the major wave-field components are reconstructed with relatively small errors in both amplitude and spatial-temporal organization.
FNO also reconstructs the global wave-field topology with high fidelity. The principal large-scale structures and most of the heterogeneous response pattern are retained, demonstrating that the Fourier representation is effective in capturing the long-range correlations induced by the extended excitation. Nevertheless, the corresponding error map contains noticeable high-frequency and spatially correlated residuals, indicating remaining inaccuracies in the detailed amplitude and phase reconstruction. LNO exhibits a comparable global reconstruction capability, although the error becomes increasingly structured in regions containing finer-scale oscillatory features. This suggests that its latent representation captures the dominant field organization but incompletely resolves the highest-frequency components of the target response.
The performance degradation becomes more evident for WNO and GNO. Both models recover the broad spatial arrangement of the wave field, but the reconstructed patterns are visibly smoother and less detailed than the ground truth. Their error maps retain a substantial portion of the fine-scale structure, indicating incomplete recovery of the oscillatory components generated by the asymmetric long-pulse excitation. In the GNO prediction, additional regularized or banded structures are visible, suggesting that the reconstructed field is biased toward a smoother low-dimensional representation of the target dynamics.
The limitations of ResUNet and DeepONet are particularly pronounced in this case. ResUNet largely suppresses the complex distributed wave response and reconstructs only a small fraction of the dominant field components. Consequently, its residual closely resembles the ground-truth wave field over large portions of the domain, demonstrating that important physical structures are missing from the prediction. DeepONet preserves somewhat more of the large-scale spatial variation, but the prediction is strongly smoothed and lacks much of the fine spatiotemporal texture present in the target. Its residual therefore remains highly structured, particularly in regions where the reference field contains rapidly varying wave patterns.
This case is especially informative because it shows that model performance is not determined solely by the ability to localize isolated wavefronts or excitation peaks. Under a $1\,\mathrm{s}$ nonuniform long-pulse excitation, the target field contains overlapping multiscale structures generated over an extended temporal interval. Accurate reconstruction therefore requires simultaneous representation of long-range correlations, local oscillatory features, and spatially varying amplitudes. The superior performance of U-NO suggests that its multiscale encoder--decoder operator structure is particularly well suited to this combination of global and local dependencies.

\begin{figure*}[t]
	\centering
	\includegraphics[width=\textwidth]{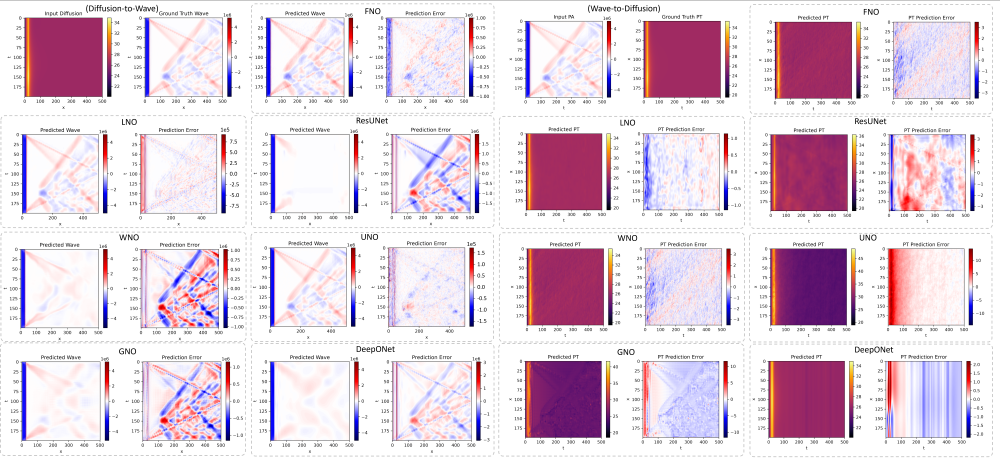}
	\caption{Qualitative comparison of diffusion-to-wave (left) and wave-to-diffusion (right) reconstruction for a plate containing multiple square defects under a uniform $1\,\mathrm{s}$ long-pulse excitation.
	The diffusion/wave input, ground-truth wave/diffusion field, and predictions obtained using FNO, LNO, ResUNet, WNO, U-NO, GNO, and DeepONet are shown together with their corresponding prediction errors.
	}
	\label{fig15}
\end{figure*}

Figure~\ref{fig15} presents a representative qualitative comparison for a plate containing multiple square defects under a uniform long-pulse excitation of $1\,\mathrm{s}$. Compared with the previously discussed single-defect or pristine cases, this configuration is substantially more challenging because the coexistence of several defects gives rise to repeated wave scattering, reflection, and mutual interaction among defect-generated wavefronts. As a result, the target wave field contains a rich superposition of coherent structures distributed over a broad spatiotemporal region. Recovering this response from the comparatively smooth diffusion input therefore requires the model to reconstruct not only the global propagation pattern, but also the complex multidefect scattering topology and its associated amplitude variations.
Among all investigated models, U-NO again provides the best qualitative agreement with the ground-truth wave field. It accurately recovers the major propagating branches, the defect-related scattered components, and the overall organization of the multiscale wave structures. Its prediction error is comparatively weak and largely unstructured, indicating that the main discrepancies are confined to small-amplitude residual fluctuations rather than systematic failure along the dominant wavefronts. This is particularly important in the present example, where errors in representing even a subset of the defect-induced scattering branches would lead to clear coherent artifacts in the residual map.
FNO also shows strong reconstruction capability, reproducing most of the global wave-field topology and the principal scattering trajectories with good fidelity. The main incident and reflected structures are well preserved, and the prediction closely follows the broad organization of the reference field. However, its residual remains more structured than that of U-NO, indicating persistent inaccuracies in the reconstruction of some local amplitudes and finer wavefront details. LNO exhibits a similar level of performance, recovering the dominant propagation and scattering structures while retaining somewhat more noticeable residuals in regions where multiple scattered components overlap. This suggests that the latent operator representation effectively captures the main cross-physics correspondence, but does not fully resolve all fine-scale interactions generated by the multidefect configuration.
The limitations of the remaining models become more apparent in this more demanding case. WNO and GNO recover only part of the target field structure, and their predictions appear visibly smoother and less detailed than the ground truth. Their error maps retain a substantial portion of the missing scattering patterns, especially in regions containing dense wavefront overlap. This indicates that these models identify the coarse wave-field organization but incompletely reproduce the detailed spatiotemporal complexity associated with multiple square defects.
ResUNet performs considerably worse, reconstructing only a limited fraction of the true response and suppressing much of the defect-related wave structure. Its residual strongly resembles the missing components of the ground-truth field, demonstrating that important physical features are not captured by the prediction. DeepONet also shows substantial smoothing and loss of detail, preserving only coarse large-scale variations while failing to recover much of the finer scattering topology. Consequently, its error map remains highly structured and contains pronounced defect-related residuals over a large portion of the domain.
This example is particularly informative because it demonstrates that the advantage of the best operator-learning models extends beyond simple propagation or isolated scattering recovery. In the multiple-defect long-pulse case, the target wave field is governed by the interaction of long-duration forcing with several spatially separated discontinuities, producing a highly nonlocal and multiscale response. Accurate reconstruction therefore requires simultaneous representation of long-range propagation, repeated scattering, and defect-interaction effects. The consistently superior performance of U-NO indicates that its multiscale operator architecture is especially well suited to such complex cross-physics mappings.

Together with the previous short-pulse, chirp, nonuniform-excitation, and long-pulse examples, the present multiple-defect case further demonstrates that the relative ranking of the models remains stable across both excitation diversity and structural complexity. In particular, U-NO consistently provides the most faithful reconstruction of global wave-field topology and local defect-induced structures, highlighting its robustness for cross-physics mapping under realistic and heterogeneous conditions.

\begin{table*}[t]
	\centering
	\caption{Performance comparison of different models for wave-to-diffusion mapping.}
	\label{tab:model_comparison2}
	\renewcommand{\arraystretch}{1.15}
	\setlength{\tabcolsep}{6pt}
	\begin{tabular}{lccccc}
		\toprule
		\textbf{Model} &
		\textbf{MSE} &
		\textbf{MAE} &
		\textbf{RMSE} &
		\textbf{Rel. $L_2$} &
		\textbf{$R^2$} \\
		\midrule
		DeepONet
		& 60.447
		& 2.411
		& 7.775
		& 0.184
		& 0.906 \\
		
		FNO
		& 58.017
		& 2.198
		& 7.617
		& 0.181
		& 0.909 \\
		
		ResUNet
		& 112.159
		& 2.982
		& 10.591
		& 0.251
		& 0.825 \\
		
		U-NO
		& 65.644
		& \textbf{2.234}
		& 8.102
		& 0.192
		& 0.898 \\
		
		WNO
		& 68.020
		& 2.402
		& 8.247
		& 0.196
		& 0.894 \\
		
		\textbf{GNO}
		& \textbf{41.903}
		& 2.355
		& \textbf{6.473}
		& \textbf{0.154}
		& \textbf{0.935} \\
		
		LNO
		& 64.377
		& 1.707
		& 8.024
		& 0.190
		& 0.899 \\
		\bottomrule
	\end{tabular}
\end{table*}

\subsection{Wave-to-Diffusion}
For the wave-to-diffusion mapping (right panels of Fig.~\ref{fig10}), all architectures exhibit stable convergence, with a rapid reduction in error during the initial training stage followed by gradual saturation. A persistent gap between the training and validation losses is observed for most models, indicating different degrees of overfitting. Among the tested architectures, LNO and FNO achieve the lowest validation errors, while WNO and U-NO also show favorable convergence. In contrast, ResUNet and GNO exhibit comparatively larger generalization gaps. The marked points denote the checkpoints selected according to the minimum validation error. Table~\ref{tab:model_comparison2} further quantifies the performance of the different architectures for the wave-to-diffusion mapping. GNO achieves the best overall accuracy, yielding the lowest MSE (41.903), RMSE (6.473), and relative $\ell_2$ error (0.154), together with the highest $R^2$ of 0.935. LNO provides the lowest MAE of 1.707, while FNO and DeepONet also exhibit competitive performance. In contrast, ResUNet shows consistently larger errors, suggesting that operator-based architectures are generally better suited to capturing the nonlocal transformation from propagative wave fields to dissipative diffusion dynamics.

The right part of Fig.~\ref{fig11} compares the wave-to-diffusion predictions obtained from different architectures. Overall, the learned models successfully recover the smooth, strongly dissipative structure of the target heat field from the oscillatory wave input, demonstrating that the wave-to-diffusion transformation is considerably more stable than the inverse mapping. Among the tested models, FNO and GNO reproduce the target field with the highest visual fidelity, leaving mainly weak and spatially localized residuals, whereas U-NO and WNO retain the dominant diffusion pattern but exhibit more pronounced structured errors. ResUNet and DeepONet show larger systematic deviations, particularly along the temporal direction, indicating a reduced ability to reconstruct the global diffusive evolution.

The right part of Fig.~\ref{fig12} shows the wave-to-diffusion predictions for a representative case containing two spatially localized features. All architectures recover the dominant diffusive structures and correctly identify their spatial positions, confirming that the propagative wave field retains sufficient information for reconstructing the corresponding heat response. FNO, LNO, WNO, U-NO, and GNO reproduce the principal thermal bands with relatively high fidelity, although their residual maps reveal different levels of localized amplitude and temporal errors. In comparison, ResUNet exhibits broader systematic deviations, while DeepONet produces more pronounced vertically structured residuals and stronger smoothing of the target field. These results further indicate that operator-based models generally provide a more faithful wave-to-diffusion reconstruction than the conventional convolutional baseline.

The right part of Fig.~\ref{fig13} presents the wave-to-diffusion reconstruction for a more complex wave response containing multiple localized features. Most models recover the principal spatial locations and the slowly varying background of the ground-truth heat field, demonstrating that the diffusive representation can be inferred even from a comparatively structured wave response. FNO and LNO provide particularly clean reconstructions, with residuals mainly confined to the vicinity of the localized thermal features, while WNO also preserves these structures but exhibits larger amplitude deviations. ResUNet produces a noticeably smoother prediction with broader residual patterns, whereas U-NO shows more spatially distributed errors. For this realization, GNO introduces visible temporally structured artifacts despite recovering the dominant features, and DeepONet tends to oversmooth the localized responses, leading to pronounced vertically correlated residuals.

The right part of Fig.~\ref{fig14} considers a challenging wave-to-diffusion case in which the wave input exhibits strong oscillatory and spatially complex wave patterns, whereas the corresponding heat field is comparatively smooth. FNO, LNO, and U-NO recover this smooth diffusive background with relatively high fidelity, with their residuals dominated by weak distributed fluctuations. WNO also captures the overall thermal level but introduces more pronounced localized artifacts, while ResUNet exhibits broad low-frequency deviations across the field. In contrast, GNO retains substantial wave-like spatial structure in its prediction, indicating incomplete suppression of propagative features, and DeepONet produces an overly smooth, vertically banded reconstruction. This example highlights that successful wave-to-diffusion mapping requires the learned operator to strongly attenuate oscillatory wave content while preserving the slowly varying thermal component.

The right part of Fig.~\ref{fig15} further demonstrates the wave-to-diffusion mapping for a wave field containing pronounced oblique and interference-like wave structures, while the corresponding heat response is nearly spatially uniform. FNO, LNO, and WNO recover the smooth thermal field with relatively small residuals, indicating effective suppression of the oscillatory wave components. U-NO also reproduces the overall diffusion pattern, although a noticeable boundary bias remains. In contrast, ResUNet and GNO retain stronger spatially structured artifacts, suggesting incomplete removal of wave-related features, whereas DeepONet produces an overly smooth prediction accompanied by pronounced vertically correlated residuals. Overall, this case again highlights the advantage of operator-based spectral and multiscale representations for extracting the slowly varying diffusive component from complex wave observations.

\section{Conclusion}
We introduced cross-physics mapping as an operator-learning framework for translating fields governed by distinct physical laws. By representing heterogeneous fields through compatible latent operators and matching their dominant dimensionless scales, the proposed framework provides a principled means of constructing paired cross-physics data without assuming analytical or dynamical equivalence between the underlying partial differential equations. Its feasibility was examined through bidirectional mappings between diffusive and wave fields using seven representative convolutional and neural-operator architectures.
The results reveal a pronounced directional asymmetry. Diffusion-to-wave mapping is substantially more challenging because it requires reconstructing sharp wavefronts, phase, and time-of-flight information that is strongly attenuated in the diffusive observations. U-NO achieves the best performance in this direction, with a relative $\ell_2$ error of $0.307$ and an $R^2$ of $0.905$, demonstrating the advantage of combining multiscale representations with nonlocal operator learning. Conversely, wave-to-diffusion mapping is more stable because it primarily requires suppressing oscillatory wave components while retaining slowly varying thermal structures. GNO provides the best aggregate accuracy in this direction, reaching a relative $\ell_2$ error of $0.154$ and an $R^2$ of $0.935$, while LNO yields the lowest MAE. Overall, neural operators consistently outperform the conventional convolutional baseline, confirming that cross-physics transformations are intrinsically nonlocal mappings between function spaces.
These findings show that deep learning can recover useful relationships between distinct physical modalities when they share a latent scene and compatible characteristic scales. This capability should nevertheless be understood as learning a conditional operator on the sampled physical manifold, rather than universally recovering information irreversibly removed by diffusion. Future work will therefore address identifiability, uncertainty quantification, physics-constrained learning, and experimental validation across broader sensing modalities.

\begin{acknowledgments}
This work was supported by the Adolf Martens Fellowship (Grant n. BAM-AMF-2025-1).
\end{acknowledgments}

\section*{Data Availability Statement}

AIP Publishing believes that all datasets underlying the conclusions of the paper should be available to readers. Authors are encouraged to deposit their datasets in publicly available repositories or present them in the main manuscript. All research articles must include a data availability statement stating where the data can be found. In this section, authors should add the respective statement from the chart below based on the availability of data in their paper.

\appendix

\section{Derivation of the Cross-Physics Scaling Principle}
\label{app:scaling}
This appendix derives the dimensionless scaling principle used to construct paired diffusion--wave data. The purpose of the scaling is not to establish an analytical equivalence between parabolic and hyperbolic partial differential equations. Instead, it provides a controlled similarity criterion under which the two physical processes probe comparable spatial scales over a common observation interval.

We begin with the homogeneous diffusion equation
\begin{equation}
	\frac{\partial u}{\partial t}
	=
	\alpha \nabla^2 u,
	\label{eq:app_diffusion}
\end{equation}
where $u(\mathbf{x},t)$ is the diffusive field,
$\mathbf{x}\in\Omega\subset\mathbb{R}^d$, and
$\alpha$ is the diffusivity with dimensions
$
	[\alpha]
	=
	L^2/T.
$
Let $L$ denote a characteristic length scale and $t_0$ a characteristic observation time. We introduce the dimensionless variables
$
	\mathbf{x}
	=
	L\mathbf{x}^*,
	t
	=
	t_0 t^*,
	u
	=
	U_0 u^*,
$
where $U_0$ is an arbitrary characteristic field amplitude.
The temporal derivative becomes
\begin{equation}
	\frac{\partial u}{\partial t}
	=
	\frac{U_0}{t_0}
	\frac{\partial u^*}{\partial t^*},
\end{equation}
whereas the Laplacian transforms as
\begin{equation}
	\nabla^2u
	=
	\frac{U_0}{L^2}
	\nabla^{*2}u^*.
\end{equation}
Substitution into Eq.~\eqref{eq:app_diffusion} gives
\begin{equation}
	\frac{U_0}{t_0}
	\frac{\partial u^*}{\partial t^*}
	=
	\alpha
	\frac{U_0}{L^2}
	\nabla^{*2}u^*.
\end{equation}
After cancellation of $U_0$,
\begin{equation}
	\frac{\partial u^*}{\partial t^*}
	=
	\underbrace{
		\frac{\alpha t_0}{L^2}
	}_{\mathrm{Fo}}
	\nabla^{*2}u^*.
	\label{eq:app_diffusion_nd}
\end{equation}

The dimensionless parameter
$
	\mathrm{Fo}
	=
	\alpha t_0/L^2
$
is the Fourier number.
Equivalently, the characteristic diffusion time over a distance $L$ is
$
	\tau_d
	=
	L^2/\alpha,
$
such that
$
	\mathrm{Fo}
	=
	t_0/\tau_d.
$
Thus, $\mathrm{Fo}$ measures the observation duration relative to the time required for diffusion to act over the characteristic length $L$.

The same scale can be obtained directly from the diffusion Green's function. For an impulsive excitation in an unbounded $d$-dimensional medium,
\begin{equation}
	G_d(\mathbf{x},t)
	=
	\frac{1}{
		(4\pi\alpha t)^{d/2}
	}
	\exp
	\left(
	-
	\frac{
		|\mathbf{x}|^2
	}{
		4\alpha t
	}
	\right).
	\label{eq:app_diff_green}
\end{equation}
The exponential term shows that the characteristic distance influenced by diffusion after a time $t$ scales as
$
	\ell_d(t)
	\sim
	\sqrt{\alpha t},
$
up to a convention-dependent numerical factor.
At $t=t_0$,
\begin{equation}
	\frac{\ell_d(t_0)}{L}
	\sim
	\sqrt{
		\frac{\alpha t_0}{L^2}
	}
	=
	\sqrt{\mathrm{Fo}}.
	\label{eq:app_diff_length_nd}
\end{equation}
Hence, $\sqrt{\mathrm{Fo}}$ has a direct geometric interpretation: it is the characteristic diffusion penetration distance normalized by the system length.
The irreversible smoothing property of diffusion can also be seen in spectral space. Taking the spatial Fourier transform of Eq.~\eqref{eq:app_diffusion} yields
\begin{equation}
	\frac{\partial \widehat{u}}{\partial t}
	=
	-
	\alpha |\mathbf{k}|^2
	\widehat{u},
\end{equation}
with solution
\begin{equation}
	\widehat{u}(\mathbf{k},t)
	=
	\exp
	\left(
	-\alpha|\mathbf{k}|^2t
	\right)
	\widehat{u}_0(\mathbf{k}).
	\label{eq:app_diff_spec}
\end{equation}
Introducing
$
	\mathbf{k}^*
	=
	L\mathbf{k},
$
the attenuation becomes
\begin{equation}
	\widehat{u}(\mathbf{k}^*,t^*)
	=
	\exp
	\left[
	-
	\mathrm{Fo}
	|\mathbf{k}^*|^2
	t^*
	\right]
	\widehat{u}_0(\mathbf{k}^*).
	\label{eq:app_diff_spec_nd}
\end{equation}
Therefore, $\mathrm{Fo}$ also controls the rate at which spatial frequencies are attenuated over the normalized observation interval.

We next consider a homogeneous wave field satisfying
\begin{equation}
	\frac{\partial^2 v}{\partial t^2}
	=
	c^2\nabla^2v,
	\label{eq:app_wave}
\end{equation}
where $c$ is the propagation speed,
$
	[c]
	=
	L/T.
$
Using
$
	\mathbf{x}
	=
	L\mathbf{x}^*,
	t
	=
	t_0t^*,
	v
	=
	V_0v^*,
$
we obtain
\begin{equation}
	\frac{V_0}{t_0^2}
	\frac{\partial^2v^*}{\partial t^{*2}}
	=
	c^2
	\frac{V_0}{L^2}
	\nabla^{*2}v^*.
\end{equation}
Therefore,
\begin{equation}
	\frac{\partial^2v^*}{\partial t^{*2}}
	=
	\underbrace{
		\left(
		\frac{ct_0}{L}
		\right)^2
	}_{\Lambda^2}
	\nabla^{*2}v^*.
	\label{eq:app_wave_nd}
\end{equation}

We define
$
	\Lambda
	=
	ct_0/L
$
as the dimensionless propagation distance.
The characteristic wave transit time across the domain is
$
	\tau_w
	=
	L/c,
$
and hence
$
	\Lambda
	=
	t_0/\tau_w.
$
Physically, a wave propagates over a distance
$
	\ell_w(t)
	=
	ct.
$
At the characteristic observation time,
\begin{equation}
	\frac{\ell_w(t_0)}{L}
	=
	\frac{ct_0}{L}
	=
	\Lambda.
	\label{eq:app_wave_length_nd}
\end{equation}

Thus, $\Lambda$ directly measures the fraction of the characteristic domain traversed by the wave during the observation interval.

Taking the spatial Fourier transform of Eq.~\eqref{eq:app_wave} gives
\begin{equation}
	\frac{\partial^2\widehat{v}}{\partial t^2}
	+
	c^2|\mathbf{k}|^2
	\widehat{v}
	=
	0.
\end{equation}
For initial conditions
$\widehat{v}(\mathbf{k},0)=\widehat{v}_0(\mathbf{k})$
and
$\partial_t\widehat{v}(\mathbf{k},0)=\widehat{w}_0(\mathbf{k})$,
the solution is
\begin{equation}
	\widehat{v}(\mathbf{k},t)
	=
	\widehat{v}_0(\mathbf{k})
	\cos
	\left(
	c|\mathbf{k}|t
	\right)
	+
	\frac{
		\widehat{w}_0(\mathbf{k})
	}{
		c|\mathbf{k}|
	}
	\sin
	\left(
	c|\mathbf{k}|t
	\right).
	\label{eq:app_wave_spec}
\end{equation}
In dimensionless variables,
$
	c|\mathbf{k}|t
	=
	\Lambda
	|\mathbf{k}^*|
	t^*.
$
Therefore, the normalized wave evolution is controlled by $\Lambda$, whereas the normalized diffusion evolution is controlled by $\mathrm{Fo}$.
The spectral structures of the two equations remain fundamentally different:
$
	e^{-\mathrm{Fo}|\mathbf{k}^*|^2t^*}
$ (diffusion),
whereas
$
	e^{\pm i\Lambda|\mathbf{k}^*|t^*}
$ (wave propagation).
The former produces irreversible attenuation, whereas the latter produces phase evolution. Consequently, matching dimensionless coefficients does not make the two evolution operators identical.

The objective of the scaling principle is to ensure that the two fields probe comparable spatial extents during the same characteristic observation interval $t_0$.
For diffusion,
$
	\ell_d(t_0)/L
	\sim
	\sqrt{\mathrm{Fo}},
$
whereas for wave propagation,
$
	\ell_w(t_0)/L
	=
	\Lambda.
$
A natural cross-physics similarity condition is therefore
$
	\ell_d(t_0)/L
	\sim
	\ell_w(t_0)/L.
$
Substitution gives
$
	\sqrt{\mathrm{Fo}}
	\sim
	\Lambda.
$
Squaring both sides yields
\begin{equation}
	\mathrm{Fo}
	\sim
	\Lambda^2.
	\label{eq:app_scaling}
\end{equation}

Using the definitions
$
	\mathrm{Fo}
	=
	\alpha t_0/L^2,
	\Lambda
	=
	ct_0/L,
$
Eq.~\eqref{eq:app_scaling} becomes
\begin{equation}
	\frac{\alpha t_0}{L^2}
	\sim
	\frac{c^2t_0^2}{L^2}.
\end{equation}
After cancellation of $L^2$ and one factor of $t_0$,
$
	\alpha
	\sim
	c^2t_0.
$
The corresponding effective propagation speed is therefore
$
	c_{\mathrm{eff}}
	\sim
	\sqrt{
		\alpha/t_0
	}.
$
This expression also follows directly by matching the characteristic distances:
$
	\sqrt{\alpha t_0}
	\sim
	c_{\mathrm{eff}}t_0,
$
which immediately gives
$
	c_{\mathrm{eff}}
	\sim
	\sqrt{
		\alpha/t_0
	}.
$

The same result can be expressed through the characteristic diffusion and wave time scales,
$
	\tau_d
	=
	L^2/\alpha,
	\tau_w
	=
	L/c.
$
Since
$
	\mathrm{Fo}
	=
	t_0/\tau_d,
	\qquad
	\Lambda
	=
	t_0/\tau_w,
$
the similarity condition
$\mathrm{Fo}=\Lambda^2$
can equivalently be written as
\begin{equation}
	\frac{t_0}{\tau_d}
	=
	\left(
	\frac{t_0}{\tau_w}
	\right)^2.
\end{equation}
Hence,
$
	\tau_w^2
	=
	t_0\tau_d.
$
Thus, the selected wave time scale is the geometric mean relation induced by matching diffusion penetration and wave travel distances over $t_0$.

Exact equality of the two characteristic distances is not required. More generally, we may introduce a dimensionless scaling factor $\chi>0$,
$
	\ell_w(t_0)
	=
	\chi
	\ell_d(t_0).
$
This gives
$
	\Lambda
	=
	\chi
	\sqrt{\mathrm{Fo}},
$
or
$
	\Lambda^2
	=
	\chi^2\mathrm{Fo}.
$
The corresponding propagation speed is
\begin{equation}
	c
	=
	\chi
	\sqrt{
		\frac{\alpha}{t_0}
	}.
	\label{eq:app_general_c}
\end{equation}
The special case $\chi=1$ corresponds to equal characteristic spatial reach and is used as the default scaling in this work.
Introducing $\chi$ is useful conceptually because it separates the physical parameters from the choice of cross-domain alignment. It also permits controlled ablation studies in which the target wave propagation range is shorter or longer than the characteristic diffusion penetration depth.

For heterogeneous fields,
\begin{equation}
	\frac{\partial u}{\partial t}
	=
	\nabla\cdot
	\left[
	\alpha(\mathbf{x})
	\nabla u
	\right],
    \quad
	\frac{\partial^2v}{\partial t^2}
	=
	\nabla\cdot
	\left[
	c^2(\mathbf{x})
	\nabla v
	\right],
\end{equation}
there is generally no single global diffusion length or wave speed. We therefore define reference background parameters
$\alpha_{\mathrm{ref}}$ and $c_{\mathrm{ref}}$ and construct
\begin{equation}
	\mathrm{Fo}_{\mathrm{ref}}
	=
	\frac{
		\alpha_{\mathrm{ref}}t_0
	}{
		L^2
	},
    \quad
	\Lambda_{\mathrm{ref}}
	=
	\frac{
		c_{\mathrm{ref}}t_0
	}{
		L
	}.
\end{equation}
The reference scaling condition is then
$
	\mathrm{Fo}_{\mathrm{ref}}
	=
	\Lambda_{\mathrm{ref}}^2.
$
Local material contrasts may subsequently be represented as
\begin{equation}
	\alpha(\mathbf{x})
	=
	\alpha_{\mathrm{ref}}
	a(\mathbf{x}),
	\qquad
	c(\mathbf{x})
	=
	c_{\mathrm{ref}}
	q(\mathbf{x}),
\end{equation}
where $a(\mathbf{x})$ and $q(\mathbf{x})$ are dimensionless coefficient fields. The global scaling fixes the overall temporal--spatial regime, whereas the heterogeneous coefficient maps determine how the shared latent structure perturbs each physical field.

The resulting paired simulations may be expressed abstractly as
\begin{equation}
	u
	=
	\mathcal{S}_{d}
	\left(
	\boldsymbol{\theta};
	\mathrm{Fo}
	\right),
    \quad
	v
	=
	\mathcal{S}_{w}
	\left(
	\boldsymbol{\theta};
	\Lambda
	\right),
\end{equation}
where $\boldsymbol{\theta}$ denotes the shared latent structural and excitation parameters.
The similarity constraint
$
	\sqrt{\mathrm{Fo}}
	\sim
	\Lambda
$
ensures that the two solution operators explore comparable normalized spatial scales. The learning problem can then be written as
$
	\mathcal{G}_{\theta}:
	\mathcal{O}_{\Gamma}
	\mathcal{S}_{d}
	\left(
	\boldsymbol{\theta}
	\right)
	\longmapsto
	\mathcal{O}_{\Gamma}
	\mathcal{S}_{w}
	\left(
	\boldsymbol{\theta}
	\right),
$
where $\mathcal{O}_{\Gamma}$ denotes the surface observation operator.
Importantly, Eq.~\eqref{eq:app_scaling} is a similarity constraint for dataset construction, not an identity between $\mathcal{S}_{d}$ and $\mathcal{S}_{w}$. In particular,
$
	\mathcal{S}_{d}
	\neq
	\mathcal{S}_{w},
$
and the corresponding spectra exhibit fundamentally different dynamics:
$
	e^{-\alpha|\mathbf{k}|^2t}
$
versus
$
	e^{\pm ic|\mathbf{k}|t}.
$
The role of the scaling is only to avoid pairing a diffusion process and a wave process whose characteristic spatial reaches differ by arbitrary orders of magnitude.

The quantity
$
	c_{\mathrm{eff}}
	=
	\sqrt{
		\alpha/t_0
	}
$
should not be interpreted as a material wave velocity derived from the diffusion equation. In particular, Fourier diffusion remains parabolic and possesses no finite physical propagation speed. Rather, $c_{\mathrm{eff}}$ is a cross-domain scaling parameter defined relative to the selected observation time $t_0$.
This dependence on $t_0$ is essential. If the observation interval is changed, the effective wave speed required to maintain equal characteristic spatial reach changes according to
$
	c_{\mathrm{eff}}
	\propto
	t_0^{-1/2}.
$
Therefore, the scaling does not convert diffusion into a physical wave process. It establishes a dimensionally consistent correspondence between the characteristic spatial extents of two distinct evolution equations.
In summary, the scaling principle used throughout this work is
\begin{equation}
	\frac{\ell_d}{L}
	\sim
	\sqrt{\mathrm{Fo}}
	\sim
	\Lambda
	\sim
	\frac{\ell_w}{L}
\end{equation}
or equivalently
$
	\mathrm{Fo}
	\sim
	\Lambda^2,
	c_{\mathrm{eff}}
	\sim
	\sqrt{
		\alpha/t_0
	}.
$
This relation provides the dimensionless similarity condition used to generate cross-physics training pairs while preserving the distinct dissipative and propagative nature of the underlying physical systems.

\section{Virtual wave transform results}
\label{virtualwave}

\begin{figure*}[t]
	\centering
	\includegraphics[width=\textwidth]{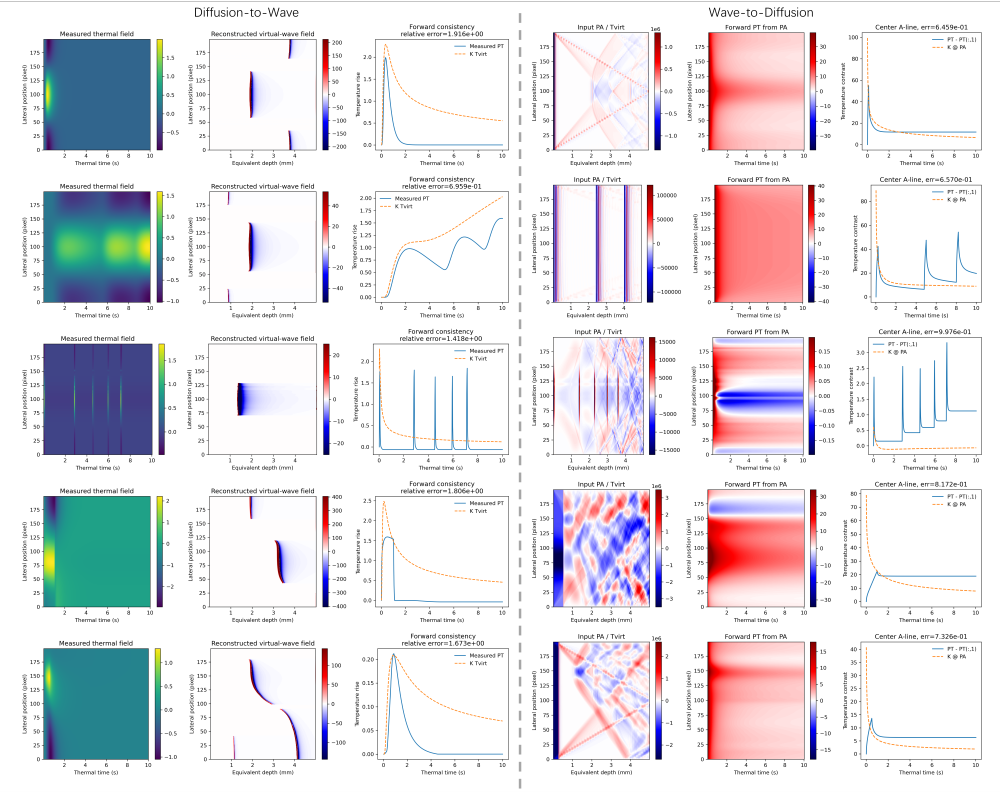}
	\caption{
		Representative results obtained using the conventional virtual-wave
		transformation under different structural and excitation conditions:
		a plate containing a square-hole defect under a $0.01\,\mathrm{s}$
		short-pulse excitation;
		a defect-free plate under chirp excitation;
		representative cases with asymmetric, spatially nonuniform localized
		Gaussian long-pulse excitation of $1\,\mathrm{s}$ duration; and
		a plate containing multiple square defects under a uniform
		$1\,\mathrm{s}$ long-pulse excitation.
	}
	\label{fig:virtual_wave}
\end{figure*}

The virtual-wave formulation provides a mathematical connection between
diffusive thermal transport and wave-like propagation~\cite{Peter2017,Vibro2026,GVWT2026,Uni2026}. Its central idea is
to represent the irreversible temporal evolution governed by the heat
diffusion equation through a local temporal transformation of an auxiliary
field that satisfies a reversible wave equation. The auxiliary field does
not represent a physically propagating thermal wave; rather, it is a
mathematical construction that permits wave-based reconstruction and
inverse methods to be applied to thermographic data.

Consider a homogeneous medium with thermal diffusivity $\alpha$. For an
impulsive excitation producing an initial temperature distribution
$T_0(\mathbf r)$, the temperature field satisfies
\begin{equation}
	\left(
	\nabla^2-\frac{1}{\alpha}\frac{\partial}{\partial t}
	\right)
	T(\mathbf r,t)
	=
	-\frac{1}{\alpha}
	T_0(\mathbf r)\delta(t).
	\label{eq:app_diffusion2}
\end{equation}

The source term is chosen such that
$
	T(\mathbf r,0^+)=T_0(\mathbf r).
$
We now introduce an auxiliary field
$T_{\mathrm{virt}}(\mathbf r,t')$, referred to as the virtual temperature
wave, satisfying
\begin{equation}
	\left(
	\nabla^2-\frac{1}{c^2}
	\frac{\partial^2}{\partial t'^2}
	\right)
	T_{\mathrm{virt}}(\mathbf r,t')
	=
	-\frac{1}{c^2}
	\frac{\partial}{\partial t'}
	\left[
	T_0(\mathbf r)\delta(t')
	\right],
	\label{eq:app_virtual_wave}
\end{equation}
where $c$ is an arbitrarily chosen virtual propagation velocity. Importantly,
$c$ does not correspond to a physical thermal propagation speed within the
Fourier heat-conduction model. It defines only the temporal--spatial scaling
of the auxiliary wave problem.
The diffusion field and the virtual-wave field therefore share the same
initial spatial distribution $T_0(\mathbf r)$, while their subsequent
evolutions are governed by fundamentally different operators: diffusion
for $T$ and reversible wave propagation for $T_{\mathrm{virt}}$.

Using the temporal Fourier-transform convention
\begin{equation}
	\widetilde{T}(\mathbf r,\omega)
	=
	\int_{-\infty}^{\infty}
	T(\mathbf r,t)e^{i\omega t}\,dt,
\end{equation}
\begin{equation}
	T(\mathbf r,t)
	=
	\frac{1}{2\pi}
	\int_{-\infty}^{\infty}
	\widetilde{T}(\mathbf r,\omega)e^{-i\omega t}\,d\omega,
\end{equation}

Eq.~\eqref{eq:app_diffusion2} becomes
\begin{equation}
	\left[
	\nabla^2-\sigma^2(\omega)
	\right]
	\widetilde{T}(\mathbf r,\omega)
	=
	-\frac{1}{\alpha}T_0(\mathbf r),
	\label{eq:app_diffusion_frequency}
\end{equation}
where the complex diffusion wavenumber is defined by
$
	\sigma^2(\omega)
	=
	-i\omega/\alpha,
$
with the square-root branch chosen consistently with spatial decay.
Similarly, Fourier transformation of the virtual-wave equation gives
\begin{equation}
	\left[
	\nabla^2+k^2(\Omega)
	\right]
	\widetilde{T}_{\mathrm{virt}}(\mathbf r,\Omega)
	=
	-\frac{i\Omega}{c^2}
	T_0(\mathbf r),
	\label{eq:app_wave_frequency}
\end{equation}
where
$
	k(\Omega)=\Omega/c.
$
The two Helmholtz-type equations can be analytically connected by
continuing the virtual-wave frequency into the complex plane. Setting
$
	\Omega=-ic\,\sigma(\omega)
$
gives
$
	k^2(\Omega)
	=
	-\sigma^2(\omega),
$
so that the spatial operators in
Eqs.~\eqref{eq:app_diffusion_frequency} and
\eqref{eq:app_wave_frequency} become identical. Matching their source
terms then yields the local spectral relation
\begin{equation}
		\widetilde{T}(\mathbf r,\omega)
		=
		\frac{c}
		{\alpha\,\sigma(\omega)}
		\widetilde{T}_{\mathrm{virt}}
		\left[
		\mathbf r,-ic\,\sigma(\omega)
		\right].
	\label{eq:app_spectral_virtual_transform}
\end{equation}

Equation~\eqref{eq:app_spectral_virtual_transform} shows that heat
diffusion can be regarded as an analytically continued representation of
an auxiliary wave field. The transformation is local in space: the
temperature and virtual-wave signals are related at the same spatial
position $\mathbf r$. Consequently, the transformation itself is
independent of whether the underlying problem is one-, two-, or
three-dimensional.

The same relation can be expressed directly in the time domain. Performing
the inverse Fourier transformation gives
\begin{equation}
	T(\mathbf r,t)
	=
	\int_{-\infty}^{\infty}
	K(t,t')
	T_{\mathrm{virt}}(\mathbf r,t')\,dt',
	\label{eq:app_kernel_transform}
\end{equation}
where
\begin{equation}
		K(t,t')
		=
		\frac{c}
		{\sqrt{\pi\alpha t}}
		\exp\left(
		-\frac{c^2t'^2}{4\alpha t}
		\right),
		\qquad t>0 .
	\label{eq:app_virtual_kernel}
\end{equation}

For a causal formulation, the integration interval may be restricted
according to the temporal support adopted for
$T_{\mathrm{virt}}$.
Equation~\eqref{eq:app_kernel_transform} is the fundamental virtual-wave
transformation. For every physical diffusion time $t$, the measured
temperature is a Gaussian-weighted superposition of the virtual-wave
signal over the auxiliary time coordinate $t'$. The width of the kernel
scales as
$
	\Delta t'
	\sim
	\sqrt{\alpha t}/c,
$
which directly reflects the increasing temporal spreading associated with
diffusion.
Upon temporal discretization, Eq.~\eqref{eq:app_kernel_transform} becomes
$
	\mathbf T
	=
	\mathbf K\mathbf T_{\mathrm{virt}},
$
where $\mathbf K$ is the discrete virtual-wave transformation matrix.
Recovering the virtual wave from measured thermographic data therefore
requires solving
$
	\mathbf T_{\mathrm{virt}}
	\simeq
	\mathbf K^\dagger\mathbf T,
$
where $\mathbf K^\dagger$ denotes a regularized inverse rather than the
ordinary matrix inverse.

The forward transformation
$\mathbf T_{\mathrm{virt}}\rightarrow\mathbf T$ is smoothing. High-frequency
components of the virtual-wave field are strongly attenuated during
diffusive transport, causing the singular values of $\mathbf K$ to decay
rapidly. Consequently, direct inversion is ill-conditioned and strongly
amplifies measurement noise. We therefore reconstruct the virtual-wave
field as the regularized inverse problem
\begin{equation}
	\mathbf T_{\mathrm{virt}}
	=
	\arg\min_{\mathbf x}
	{\frac{1}{2}
		||\mathbf K\mathbf x-\mathbf T||_2^2
		+
		\lambda\mathcal R(\mathbf x)},
	\label{eq:app_regularized_inverse}
\end{equation}
where $\mathcal R$ incorporates prior information, such as sparsity,
piecewise smoothness, nonnegativity, or other structural constraints on
the virtual-wave field. To solve Eq.~\eqref{eq:app_regularized_inverse}
using the alternating direction method of multipliers (ADMM), an auxiliary
variable $\mathbf z$ is introduced:
\begin{equation}
	\min_{\mathbf x,\mathbf z}
	\frac{1}{2}||\mathbf K\mathbf x-\mathbf T||_2^2
	+
	\lambda\mathcal R(\mathbf z),
	\text{subject to }
	\mathbf x=\mathbf z .
	\label{eq:app_admm_split}
\end{equation}
Using the scaled dual variable $\mathbf d$ and the penalty parameter
$\rho>0$, the ADMM iterations are
\begin{equation}
	\begin{split}
	\mathbf x^{k+1}
	&=
	\left(
	\mathbf K^{\mathsf T}\mathbf K+\rho\mathbf I
	\right)^{-1}
	\left[
	\mathbf K^{\mathsf T}\mathbf T
	+
	\rho\left(\mathbf z^{k}-\mathbf d^{k}\right)
	\right],
	\\
	\mathbf z^{k+1}
	&=
	\operatorname{prox}_{(\lambda/\rho)\mathcal R}
	\left(
	\mathbf x^{k+1}+\mathbf d^{k}
	\right),
	\\
	\mathbf d^{k+1}
	&=
	\mathbf d^{k}
	+
	\mathbf x^{k+1}
	-
	\mathbf z^{k+1},
	\end{split}
\end{equation}
where
\begin{equation}
	\operatorname{prox}_{\gamma\mathcal R}(\mathbf q)
	=
	\arg\min_{\mathbf z}{
		\gamma\mathcal R(\mathbf z)
		+
		\frac{1}{2}||\mathbf z-\mathbf q||_2^2}
\end{equation}
is the proximal operator associated with the selected regularizer. The
iterations are terminated when both the primal residual
$\|\mathbf x^{k}-\mathbf z^{k}\|_2$ and the dual residual
$\rho\|\mathbf z^{k}-\mathbf z^{k-1}\|_2$ fall below prescribed
tolerances. Here, $\lambda$ determines the balance between data fidelity
and prior regularization, whereas $\rho$ primarily controls the numerical
conditioning and convergence of ADMM. Thus, ADMM provides a stable and
flexible computational framework for the inverse transformation, but it
does not eliminate the underlying ill-posedness. The need for
regularization ultimately reflects the irreversible loss of recoverable
high-frequency information caused by diffusive transport, rather than a
limitation introduced by the virtual-wave representation itself.

Once $T_{\mathrm{virt}}$ has been reconstructed, its evolution is governed
by the wave equation rather than the diffusion equation. Standard
wave-based reconstruction techniques, including time reversal,
back-projection, synthetic-aperture focusing, or other wave-field inverse
methods, can therefore be applied to the recovered virtual signal.
The complete reconstruction chain may consequently be expressed as
\begin{equation}
	T(\mathbf r_s,t)
	\xrightarrow[\mathrm{regularized}]
	{\mathbf K^{-1}}
	T_{\mathrm{virt}}(\mathbf r_s,t')
	\xrightarrow{\text{wave reconstruction}}
	T_0(\mathbf r),
	\label{eq:app_reconstruction_chain}
\end{equation}
where $\mathbf r_s$ denotes the measurement surface. This decomposition
separates the irreversible diffusive component of the problem from the
geometrical wave-propagation component: the first stage compensates for
diffusive temporal spreading, whereas the second stage exploits the
reversible propagation structure of the virtual field.

To first assess whether the diffusion-to-wave correspondence can be
recovered by a conventional physics-based transformation, we applied the
virtual-wave transform to representative samples covering different defect
configurations and excitation conditions. The results are shown in
Fig.~\ref{fig:virtual_wave}. Overall, the conventional virtual-wave
reconstruction exhibits only limited agreement with the corresponding
wave-domain fields. Although several dominant spatial features remain
recognizable, the reconstructed fields show pronounced amplitude
distortion, loss of oscillatory structure, spurious background components,
and strong sensitivity to the excitation waveform.

Figure~\ref{fig:virtual_wave} highlights the intrinsic asymmetry of the virtual-wave transformation. In the diffusion-to-wave direction (left), the inverse operator converts the smooth thermal measurements into sharply localized wavefronts, thereby recovering spatial interfaces and transient features that are strongly blurred by diffusion. This apparent resolution enhancement, however, is accompanied by substantial amplitude amplification and imperfect forward consistency, with relative errors of approximately $0.96\!-\!1.92$, reflecting the severe ill-conditioning of the inverse diffusion problem and its sensitivity to components outside the range of the forward operator. Conversely, the wave-to-diffusion mapping (right) acts as a strongly dissipative low-pass operator: oscillatory wavefronts, reflections, and interference patterns are largely suppressed and replaced by broad, slowly varying thermal distributions. Nevertheless, residual structures and centerline discrepancies remain, particularly for impulsive or highly oscillatory inputs, yielding relative errors of $0.65\!-\!1.00$. These results demonstrate that the virtual-wave correspondence is not an exact bidirectional equivalence for arbitrary fields: diffusion-to-wave reconstruction requires regularized analytic continuation, whereas wave-to-diffusion is stable but many-to-one and produces a physically consistent thermal field only when the input wavefield belongs to the admissible range of the virtual-wave model.

In particular, these results reveal an important limitation of applying the classical
virtual-wave transform as a direct diffusion-to-wave mapping. The transform
is derived most naturally for an impulsively generated initial temperature
distribution, whereas the present dataset deliberately contains finite-width
pulses, chirped excitations, long-duration heating, spatially nonuniform
sources, and heterogeneous structures. Under these conditions, the measured
diffusion field contains source-history and structural contributions that
cannot, in general, be represented by the ideal virtual-wave kernel alone.
Moreover, inversion of the diffusion operator is intrinsically
ill-conditioned because the high-frequency information required to recover
sharp propagating features has already been strongly attenuated by thermal
diffusion. Consequently, regularization suppresses precisely those spectral
components that are important for reconstructing the oscillatory wave field.

The poor performance of the analytical baseline therefore does not imply
that the two physical domains are unrelated. Rather, it demonstrates that
their correspondence cannot generally be reduced to the idealized
closed-form virtual-wave transformation. This observation motivates the
operator-learning formulation adopted here: instead of assuming an exact
analytical equivalence between the diffusion and wave equations, the neural
operator learns the more general mapping conditioned on the common latent
structure, excitation, and dimensionless physical scales represented in the
paired dataset.

\nocite{*}
\bibliography{references}

\end{document}